%% file: paper.tex
\documentclass[10pt,twocolumn,letterpaper]{article}

\usepackage[final]{cvpr}      %

\input{resources/preamble}

\definecolor{cvprblue}{rgb}{0.21,0.49,0.74}
\usepackage[pagebackref,breaklinks,colorlinks,allcolors=cvprblue]{hyperref}

\def\paperID{*****} %
\def\confName{CVPR}
\def\confYear{2026}

\title{Rethinking UMM Visual Generation: \\ Masked Modeling for Efficient Image-Only Pre-training}

\author{Peng Sun$^{1,3,}\thanks{Equal Contritbution}$ \quad Jun Xie$^{1,2,3,*}$ \quad Tao Lin$^{3,}\thanks{Corresponding author.}$ \\
\quad $^{1}$Zhejiang University \quad $^{2}$Shanghai Innovation Institute \quad $^{3}$Westlake University \\
{\tt\small sunpeng@westlake.edu.cn, junxiecs@zju.edu.cn, lintao@westlake.edu.cn} \\
}

\begin{document}
\maketitle
\input{resources/main}

\section*{Acknowledgement}
This work was supported in part by the National Science and Technology Major Project (No.\ 2022ZD0115101), NSFC under No.\ 62576285, the Research Center for Industries of the Future (RCIF) at Westlake University, and the Westlake Education Foundation.
    {
        \small
        \bibliographystyle{ieeenat_fullname}
        \bibliography{resources/reference}
    }

\clearpage
\appendix
\input{resources/appendix.tex}

\end{document}

%% file: resources/preamble.tex
\usepackage{amsmath,amssymb,amsthm}

\providecommand{\norm}[1]{\left\lVert#1\right\rVert}

\providecommand{\EEb}[2]{{\mathbb E}_{#1}\left[#2\right] } %

\providecommand{\dm}{\mathrm{d}}

\providecommand{\0}{\mathbf{0}}

\providecommand{\cc}{\mathbf{c}}

\providecommand{\hh}{\mathbf{h}}

\providecommand{\tt}{\mathbf{t}}

\providecommand{\xx}{\mathbf{x}}

\providecommand{\zz}{\mathbf{z}}

\providecommand{\mI}{\mathbf{I}}

\providecommand{\mtheta}{\boldsymbol{\theta}}

\providecommand{\mv}{\boldsymbol{v}}
\providecommand{\mq}{\boldsymbol{q}}
\providecommand{\mg}{\boldsymbol{g}}
\providecommand{\mmF}{\boldsymbol{F}}

\providecommand{\cN}{\mathcal{N}}

\newenvironment{talign*}
{\csname align*\endcsname}
{\endalign}

\usepackage[utf8]{inputenc}         %
\usepackage[T1]{fontenc}            %
\usepackage{url}                    %
\usepackage{booktabs}               %
\usepackage{amsfonts}               %
\usepackage{nicefrac}               %
\usepackage{microtype}              %
\usepackage{xcolor}                 %
\usepackage{algorithm}
\usepackage{algorithmic}
\usepackage{graphicx}
\usepackage{subcaption}
\usepackage[flushleft]{threeparttable}
\usepackage{float}
\usepackage{multirow}
\usepackage{xspace}
\usepackage{enumitem}
\usepackage[font=small]{caption}
\usepackage{autobreak}
\usepackage{sidecap}
\usepackage{wrapfig}
\usepackage{bbding}
\usepackage[toc, page, header]{appendix}
\usepackage{tikz}
\usepackage{xcolor}
\usepackage{pifont}
\usepackage{mdframed}
\usepackage{colortbl}

\definecolor{coral}{RGB}{255,127,80}
\definecolor{darkgreen}{RGB}{0,100,0}
\definecolor{darkyellow}{RGB}{204,153,0}
\definecolor{salmon}{RGB}{250,128,114}
\definecolor{darkred}{RGB}{150,0,0}
\newcommand{\darkredtext}[1]{{\color{darkred}#1}}

\newcommand{\secref}[1]{\hyperref[#1]{\darkredtext{Sec.~\ref*{#1}}}}
\newcommand{\thmref}[1]{\hyperref[#1]{\darkredtext{Thm.~\ref*{#1}}}}
\newcommand{\defref}[1]{\hyperref[#1]{\darkredtext{Def.~\ref*{#1}}}}
\newcommand{\propref}[1]{\hyperref[#1]{\darkredtext{Prop.~\ref*{#1}}}}
\newcommand{\assumpref}[1]{\hyperref[#1]{\darkredtext{Assump.~\ref*{#1}}}}
\newcommand{\remarkref}[1]{\hyperref[#1]{\darkredtext{Rem.~\ref*{#1}}}}
\newcommand{\hypref}[1]{\hyperref[#1]{\darkredtext{Hyp.~\ref*{#1}}}}
\newcommand{\conjref}[1]{\hyperref[#1]{\darkredtext{Conj.~\ref*{#1}}}}
\newcommand{\lemref}[1]{\hyperref[#1]{\darkredtext{Lem.~\ref*{#1}}}}
\newcommand{\corref}[1]{\hyperref[#1]{\darkredtext{Cor.~\ref*{#1}}}}
\newcommand{\noteref}[1]{\hyperref[#1]{\darkredtext{Nota.~\ref*{#1}}}}
\newcommand{\claimref}[1]{\hyperref[#1]{\darkredtext{Clm.~\ref*{#1}}}}
\newcommand{\obsref}[1]{\hyperref[#1]{\darkredtext{Obs.~\ref*{#1}}}}
\newcommand{\algref}[1]{\hyperref[#1]{\darkredtext{Alg.~\ref*{#1}}}}
\newcommand{\figref}[1]{\hyperref[#1]{\darkredtext{Fig.~\ref*{#1}}}}
\newcommand{\tabref}[1]{\hyperref[#1]{\darkredtext{Tab.~\ref*{#1}}}}
\newcommand{\appref}[1]{\hyperref[#1]{\darkredtext{App.~\ref*{#1}}}}

\newtheoremstyle{custom}
{1pt} %
{1pt} %
{\itshape} %
{} %
{\bfseries} %
{} %
{ } %
{\thmname{#1} \thmnumber{#2} \thmnote{(#3)} . } %

\theoremstyle{custom}

\newtheorem{innerdefinition}{Definition}
\newtheorem{innerproposition}{Proposition}
\newtheorem{innerassumption}{Assumption}
\newtheorem{innerremark}{Remark}
\newtheorem{innertheorem}{Theorem}
\newtheorem{innerhypothesis}{Hypothesis}
\newtheorem{innerconjecture}{Conjecture}
\newtheorem{innerlemma}{Lemma}
\newtheorem{innercorollary}{Corollary}
\newtheorem{innernotation}{Notation}
\newtheorem{innerclaim}{Claim}
\newtheorem{innerproblem}{Problem}
\newtheorem*{innernproblem}{Problem}

\newtheorem{innerobservation}{Observation}

\newmdenv[
    backgroundcolor=gray!10,
    linecolor=gray!100,
    linewidth=0.8pt,
    skipabove=2pt,
    skipbelow=2pt,
    innertopmargin=10pt,
    innerbottommargin=5pt,
    innerleftmargin=5pt,
    innerrightmargin=5pt,
]{definitionframe}

\newmdenv[
    backgroundcolor=blue!10,
    linecolor=blue!100,
    linewidth=0.8pt,
    skipabove=2pt,
    skipbelow=2pt,
    innertopmargin=10pt,
    innerbottommargin=5pt,
    innerleftmargin=5pt,
    innerrightmargin=5pt,
]{propositionframe}

\newmdenv[
    backgroundcolor=green!10,
    linecolor=green!100,
    linewidth=0.8pt,
    skipabove=2pt,
    skipbelow=2pt,
    innertopmargin=10pt,
    innerbottommargin=5pt,
    innerleftmargin=5pt,
    innerrightmargin=5pt,
]{assumptionframe}

\newmdenv[
    backgroundcolor=yellow!10,
    linecolor=yellow!100,
    linewidth=0.8pt,
    skipabove=2pt,
    skipbelow=2pt,
    innertopmargin=10pt,
    innerbottommargin=5pt,
    innerleftmargin=5pt,
    innerrightmargin=5pt,
]{remarkframe}

\newmdenv[
    backgroundcolor=red!10,
    linecolor=red!100,
    linewidth=0.8pt,
    skipabove=2pt,
    skipbelow=2pt,
    innertopmargin=10pt,
    innerbottommargin=5pt,
    innerleftmargin=5pt,
    innerrightmargin=5pt,
]{theoremframe}

\newmdenv[
    backgroundcolor=purple!10,
    linecolor=purple!100,
    linewidth=0.8pt,
    skipabove=2pt,
    skipbelow=2pt,
    innertopmargin=10pt,
    innerbottommargin=5pt,
    innerleftmargin=5pt,
    innerrightmargin=5pt,
]{hypothesisframe}

\newmdenv[
    backgroundcolor=orange!10,
    linecolor=orange!100,
    linewidth=0.8pt,
    skipabove=2pt,
    skipbelow=2pt,
    innertopmargin=10pt,
    innerbottommargin=5pt,
    innerleftmargin=5pt,
    innerrightmargin=5pt,
]{conjectureframe}

\newmdenv[
    backgroundcolor=cyan!10,
    linecolor=cyan!100,
    linewidth=0.8pt,
    skipabove=2pt,
    skipbelow=2pt,
    innertopmargin=10pt,
    innerbottommargin=5pt,
    innerleftmargin=5pt,
    innerrightmargin=5pt,
]{lemmaframe}

\newmdenv[
    backgroundcolor=magenta!10,
    linecolor=magenta!100,
    linewidth=0.8pt,
    skipabove=2pt,
    skipbelow=2pt,
    innertopmargin=10pt,
    innerbottommargin=5pt,
    innerleftmargin=5pt,
    innerrightmargin=5pt,
]{corollaryframe}

\newmdenv[
    backgroundcolor=pink!10,
    linecolor=pink!100,
    linewidth=0.8pt,
    skipabove=2pt,
    skipbelow=2pt,
    innertopmargin=10pt,
    innerbottommargin=5pt,
    innerleftmargin=5pt,
    innerrightmargin=5pt,
]{notationframe}

\newmdenv[
    backgroundcolor=violet!10,
    linecolor=violet!100,
    linewidth=0.8pt,
    skipabove=2pt,
    skipbelow=2pt,
    innertopmargin=10pt,
    innerbottommargin=5pt,
    innerleftmargin=5pt,
    innerrightmargin=5pt,
]{claimframe}

\newmdenv[
    backgroundcolor=salmon!10,
    linecolor=salmon!100,
    linewidth=0.8pt,
    skipabove=2pt,
    skipbelow=2pt,
    innertopmargin=10pt,
    innerbottommargin=5pt,
    innerleftmargin=5pt,
    innerrightmargin=5pt,
]{problemframe}

\newmdenv[
    backgroundcolor=lavender!10,
    linecolor=lavender!100,
    linewidth=0.8pt,
    skipabove=2pt,
    skipbelow=2pt,
    innertopmargin=10pt,
    innerbottommargin=5pt,
    innerleftmargin=5pt,
    innerrightmargin=5pt,
]{observationframe}

\newcommand{\method}{\textsc{IOMM}\xspace} %

\newcommand{\dg}[1]{\textcolor{red}{$^{\downarrow#1}$}}
\newcommand{\up}[1]{\textcolor{darkgreen}{$^{\uparrow#1}$}}
\newcommand{\fd}[1]{\textcolor{gray!50}{#1}}
\newcommand{\rp}{$\boldsymbol{\color{red!70!black}\oplus}$}

%% file: resources/main.tex
\begin{abstract}
    Unified Multimodal Models (UMMs) are often constrained by the pre-training of their \textbf{visual generation components}, which typically relies on inefficient paradigms and scarce, high-quality text-image paired data. In this paper, we systematically analyze pre-training recipes for \textbf{UMM visual generation} and identify these two issues as the major bottlenecks.
    To address them, we propose \textbf{Image-Only Training for UMMs (\method)}, a data-efficient two-stage training framework.
    The first stage pre-trains the visual generative component \textbf{exclusively} using abundant unlabeled image-only data, thereby removing the dependency on paired data \textbf{for this costly phase}. The second stage fine-tunes the model using a mixture of unlabeled images and a small curated set of text-image pairs, leading to improved instruction alignment and generative quality.
    Extensive experiments show that \method not only improves training efficiency but also achieves state-of-the-art (SOTA) performance.
    For example, our \method-B (3.6B) model was trained from scratch using only $\sim$\textbf{1050} H800 GPU hours (with the vast majority, \textbf{1000} hours, dedicated to the efficient \textbf{image-only pre-training stage}). It achieves \textbf{0.89} on GenEval and \textbf{0.55} on WISE\textemdash surpassing strong baselines such as BAGEL-7B (0.82 \& 0.55) and BLIP3-o-4B (0.84 \& 0.50).
    Code is available \url{https://github.com/LINs-lab/IOMM}.
    \looseness=-1
\end{abstract}

\setlength{\parskip}{4pt plus4pt minus0pt}

\begin{figure*}[!t]
    \centering
    \begin{subfigure}[b]{0.99\textwidth}
        \centering
        \includegraphics[width=\linewidth]{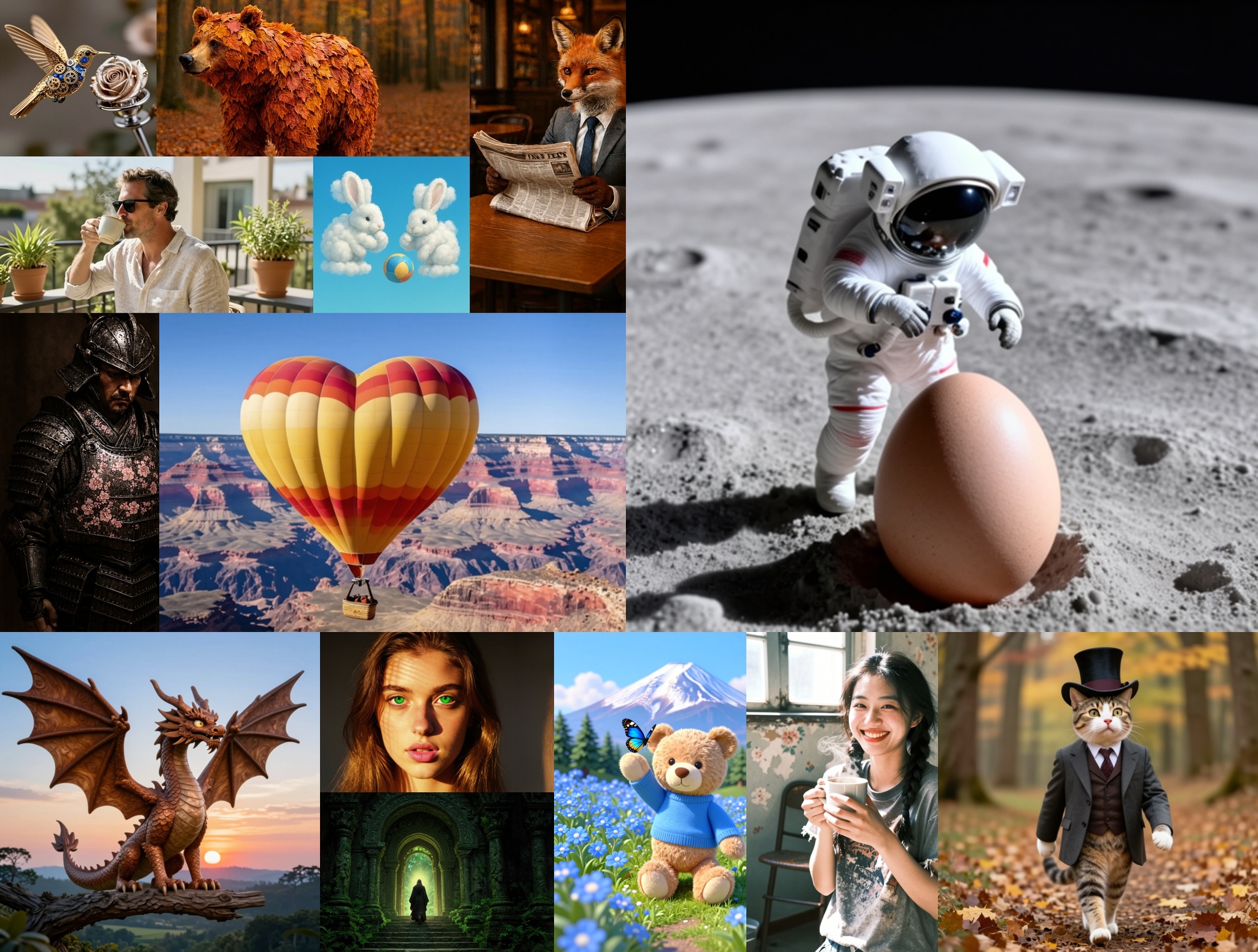}
        \caption{\textbf{Multi-resolution visualizations from our \method-XL.}}
        \label{fig:main_image}
    \end{subfigure}
    \begin{subfigure}[b]{0.59\textwidth}
        \centering
        \includegraphics[width=\linewidth]{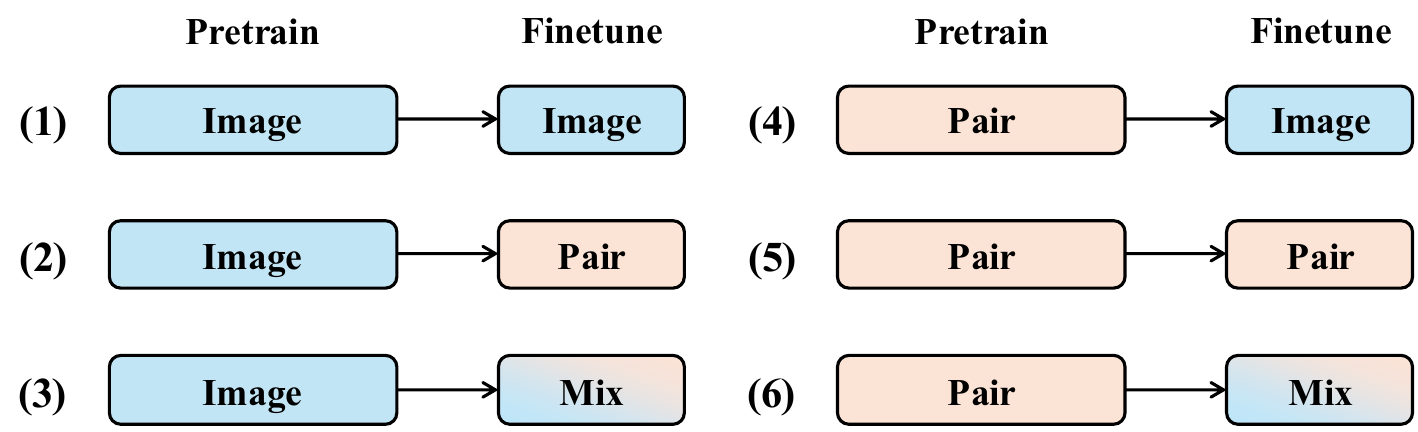}
        \caption{\textbf{Overview of training recipes.}}
        \label{fig:training_recipe}
    \end{subfigure}
    \hfill
    \begin{subfigure}[b]{0.40\textwidth}
        \centering
        \includegraphics[width=\textwidth]{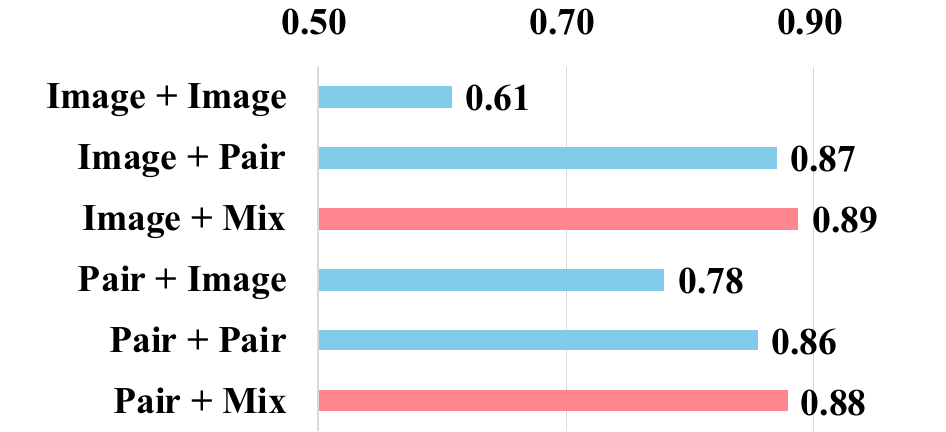}
        \caption{\textbf{GenEval performance comparison.}}
        \label{fig:6_paradigm_result}
    \end{subfigure}
    \vspace{-1em}
    \caption{\small{
            \textbf{An overview and validation of our proposed training paradigm.}
            \textbf{(a)} Visual results of our \method-XL, demonstrating high-quality, multi-resolution image synthesis. Corresponding prompts are provided in~\appref{app:prompts_details}.
            \textbf{(b)} An illustration of the six training recipes we investigate.
            \textbf{(c)} Quantitative results of six training recipes on the GenEval benchmark.
        }}
    \vspace{-1em}
\end{figure*}

\section{Introduction}
\label{sec:intro}

Unifying deep semantic understanding with rich perceptual generation in a single model is a grand challenge in AI. These UMMs promise a synergy where comprehension and generation mutually enhance one another, unlocking applications from nuanced, dialogue-based image editing to context-aware content creation~\citep{Google2025Gemini,Google2025Gemini25FlashImage,OpenAI2025Introducing}. While recent UMMs demonstrate impressive generative capabilities~\citep{wu2025qwenimagetechnicalreport,chen2025blipo,pan2025transfer,dong2024dreamllm}, their development is often hampered by significant practical constraints.

However,
current UMM training paradigms rely on vast, often proprietary, text-image datasets~\citep{chen2025blipo}. The prohibitive cost of curating this data impedes open and reproducible research. Moreover, the training procedures are notoriously inefficient, demanding immense computational resources. This raises a critical question: \textit{Can we develop a more data- and compute-efficient training paradigm for UMMs that reduces reliance on paired data while improving performance?}

In this work, we address this question by deconstructing the pre-training of UMMs' visual generative components. Our analysis reveals two primary bottlenecks: the dependency on scarce text-image pairs and the inefficiency of prevailing training objectives. We observe that many UMMs, particularly when fine-tuned on limited data, struggle to generate images that faithfully align with textual prompts. As shown in \figref{fig:raw_qwen_image_generation}, even a strong baseline like Qwen-Image~\citep{wu2025qwenimagetechnicalreport} can produce outputs that lack detail and fidelity to the input prompt.

To surmount these limitations, we introduce \method, a novel, data-efficient two-stage training paradigm for constructing and refining UMMs. Our approach commences with an unsupervised pre-training phase that leverages unlabeled, image-only data, followed by a fine-tuning stage that employs a strategic mixture of image-only and high-quality paired data. This paradigm, as we empirically demonstrate, not only mitigates the reliance on paired data but also yields superior generative quality and instruction-following capabilities.
\textbf{In summary, our contributions are threefold:}
\begin{enumerate}[label=(\alph*), nosep, leftmargin=*, topsep=3pt, itemsep=3pt]
    \item We introduce \method, a data- and compute-efficient framework built upon two key technical innovations: (1) a novel \emph{\textbf{residual query adapter}} that efficiently adapts frozen Multimodal Large Language Models (MLLMs) for generative tasks with minimal parameter overhead, and (2) a \emph{\textbf{masked image modeling}} objective that fosters a robust visual prior by framing pre-training as a sparse-to-dense reconstruction task.
    \item We present a systematic analysis of six distinct training recipes for UMMs, exploring various combinations of image-only, text-image pair, and mixed data across pre-training and fine-tuning. Under our framework \method, our central finding is that a two-stage paradigm---pre-training on image-only data followed by fine-tuning on a mixed dataset\footnote{
              Concurrent work~\citep{xie2025reconstruction} explores a similar fine-tuning strategy on mixed data, but differs crucially: (1) they focus only on fine-tuning, while we study both pre-training and fine-tuning; (2) they use standard reconstruction, whereas we use masked image modeling; (3) they test on smaller models (e.g., BAGEL-7B), while we validate on both small and large-scale UMMs (e.g., Qwen-Image-20B).
          }---yields best performance (\figref{fig:6_paradigm_result}).
    \item Extensive experiments validate the efficacy and efficiency of \method. Our resulting models attain SOTA or comparable performance across diverse benchmarks, all while operating with substantially greater data and compute efficiency (see \secref{sec:exp}).
\end{enumerate}
Additionally, we establish that our proposed \emph{\textbf{mixed-data fine-tuning strategy}} is a generalizable and effective technique for enhancing the instruction-following fidelity and image generation quality of existing powerful UMMs, which we validate on diverse models including Qwen-Image (\secref{sec:different_paradigms}).

\begin{figure*}[t]
    \centering
    \begin{subfigure}[b]{0.66\textwidth}
        \centering
        \includegraphics[width=\linewidth]{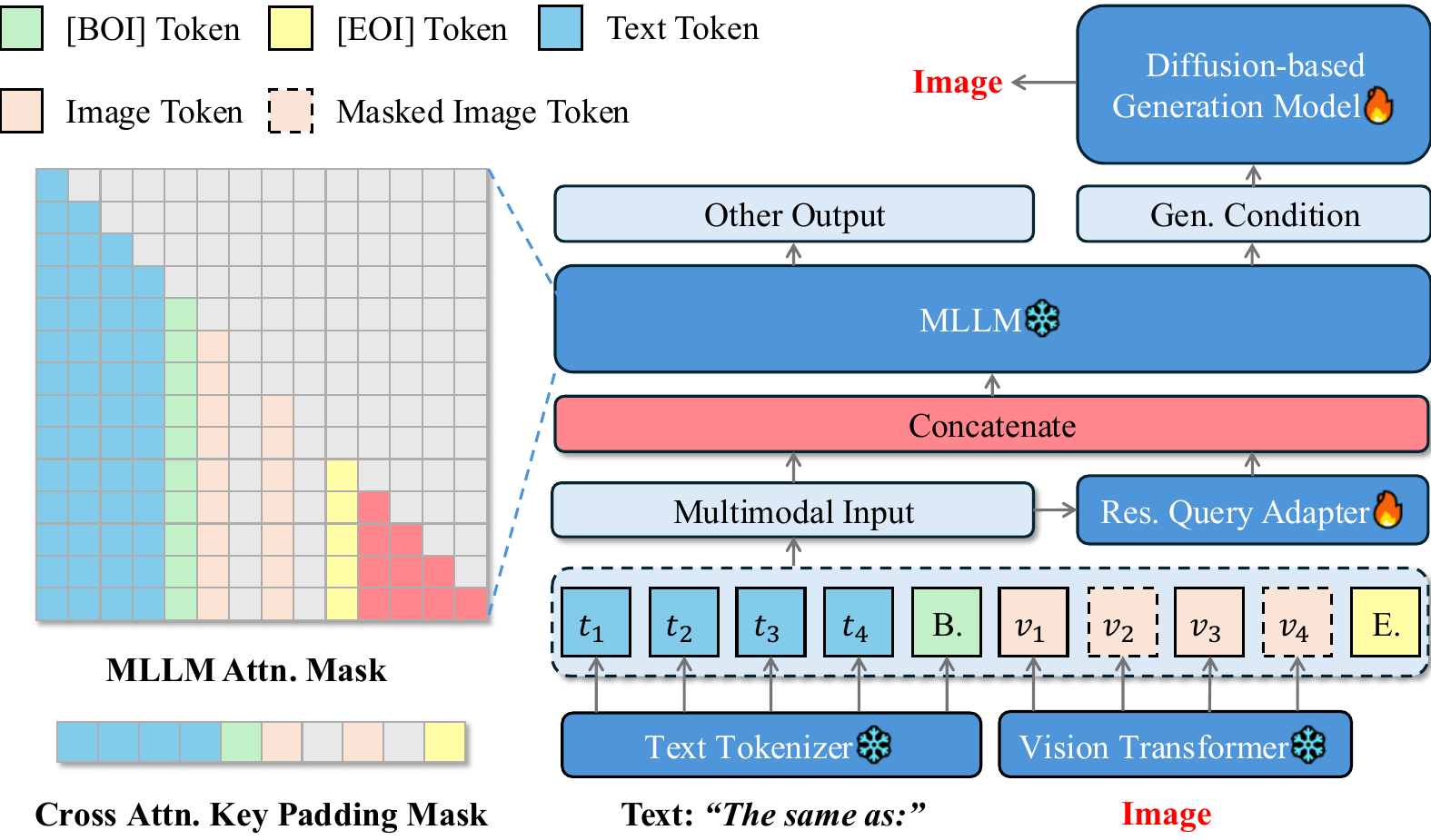}
        \caption{\textbf{The architecture of our image-only pre-training stage.}}
        \label{fig:architecture}
    \end{subfigure}
    \hfill
    \begin{subfigure}[b]{0.33\textwidth}
        \centering
        \includegraphics[width=\linewidth]{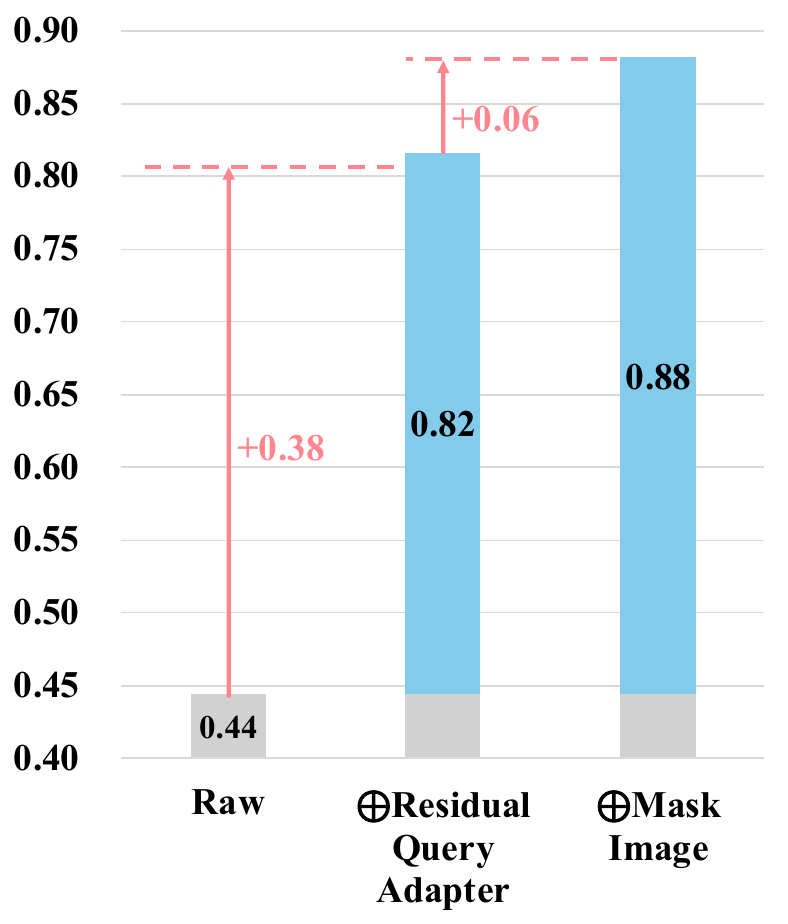}
        \caption{\textbf{Component ablation study.}}
        \label{fig:setting_performance}
    \end{subfigure}
    \vspace{-1em}
    \caption{\small{
            \textbf{Visualization of the \method framework.}
            \textbf{(a)} The architecture of our proposed framework.
            \textbf{(b)} Ablation study demonstrating the effectiveness of architectural design choices, confirming that each component contributes positively to the final GenEval score. All variants utilize the same \method-XL architecture.
        }}
    \label{fig:framework}
    \vspace{-1em}
\end{figure*}

\section{Related Work}
\label{sec:related_work}

\paragraph{Text-to-image diffusion models.}
The field of text-to-image synthesis has seen rapid advancements, driven by innovations in diffusion model architectures and training methodologies.
Foundational works, such as the initial Stable Diffusion series~\citep{rombach2022high, podell2024sdxl}, established the Latent Diffusion Model (LDM) as a dominant paradigm.
A significant architectural evolution arrived with Stable Diffusion 3~\citep{esser2024scaling}, which introduced the Multimodal Diffusion Transformer (MM-DiT). This architecture employs separate transformer-based pathways to process image and text representations independently before fusing them, markedly improving text-image alignment. Following a similar design philosophy, FLUX.1~\citep{labs2025kontext} also utilizes a dual-stream transformer architecture to enhance modality-specific encoding.

Concurrently, a parallel line of research has focused on optimizing training efficiency and data curation. For example, PixArt-$\alpha$/$\sigma$~\citep{chen2024pixart, chen2024pixarts} demonstrated the ability to achieve SOTA performance with substantially reduced training costs. Similarly, Playground v2/v2.5~\citep{playground-v2, li2024playground} is distinguished by its high aesthetic quality, a result of meticulous data filtering and reinforcement learning from user preferences. More recent models, including SANA~\citep{xie2024sana} and SANA-sprint~\citep{chen2025sana}, continue this trajectory, pushing the boundaries of performance through further architectural and training refinements.
Notably, Lumos-T2I~\citep{ma2025learning} presents a paradigm shift by demonstrating that high-quality text-to-image generation can be achieved through image-only pre-training, challenging the conventional reliance on paired text-image datasets.

However, these models are specialized for unidirectional text-to-image generation. They lack the inherent capacity for multimodal understanding, which precludes their direct application to complex, interactive tasks such as dialogue-based image editing~\citep{wu2025qwenimagetechnicalreport,Google2025Gemini} that require a seamless blend of comprehension and generation.

\paragraph{Unified understanding and generation models.}
The pursuit of models that unify multimodal understanding and generation has led to two primary training paradigms: training end-to-end from scratch, and building upon pre-trained foundation models.
Among those trained from scratch are Chameleon~\citep{team2024chameleon}, Show-o~\citep{xie2025showo}, VILA-U~\citep{wu2025vilau}, Janus~\citep{wu2024janus}, JanusPro~\citep{chen2025januspro}, JanusFlow~\citep{ma2024janusflow}, Transfusion~\citep{zhou2025transfusion}, and Harmon~\citep{wu2025harmonizing}.
These systems employ diverse architectures, including autoregressive (AR) and masked autoregressive (MAR) frameworks, to jointly handle both modalities.

The second paradigm leverages pre-trained components, integrating powerful Multimodal Large Language Models (MLLMs) with established diffusion backbones.
Notable examples include DreamLLM~\citep{dong2024dreamllm}, MetaQueries~\citep{pan2025transfer}, BLIP3-o~\citep{chen2025blipo}, UniWorld-V1~\citep{lin2025uniworldv}, Qwen-Image~\citep{wu2025qwenimagetechnicalreport}, and Bagel~\citep{deng2025emerging}. These approaches typically bridge the frozen MLLM and diffusion model using mechanisms like learnable queries or multi-stage training protocols~\citep{pan2025transfer} to harmonize understanding and generative processes. The resulting synergy of generation and comprehension enables these unified models to tackle a wide spectrum of tasks, including high-fidelity, instruction-guided image editing~\citep{Google2025Gemini,Google2025Gemini25FlashImage}.

Concurrently, UAE~\citep{yan2025unified} and ViLex~\citep{wang2025visual} explore modeling UMMs as auto-encoding tasks, which involve reconstructing the input image itself for improving understanding and generation in UMMs.

Despite these significant advances, a fundamental limitation persists across existing unified models. Current training paradigms depend heavily on meticulously curated, large-scale datasets of high-quality image-text pairs to train their generative modules. This reliance on proprietary or difficult-to-acquire data poses a significant barrier to open research and broader community-driven development.

\paragraph{Masked signal modeling.}
Masked signal modeling, pioneered by Masked Autoencoders (MAE)~\citep{he2022masked}, has become a powerful self-supervised learning paradigm. The core principle involves training a model to learn robust representations by reconstructing randomly masked portions of an input signal. Initially applied to images, this ``mask-and-predict'' strategy has been successfully adapted to a diverse range of generative tasks. Notable adaptations include predicting masked visual tokens for non-autoregressive image synthesis~\citep{chang2022maskgit}, masking textual conditions to refine guidance in diffusion models~\citep{zhou2023maskdiffusion}, leveraging attention mechanisms to generate precise editing masks from user intent~\citep{zou2024towards}, and improving the data efficiency of Generative Adversarial Network (GAN) training~\citep{huang2022masked}. The versatility of this approach underscores its potential as a flexible and potent tool for representation learning and generative modeling.

\section{Methodology}
\label{sec:methodology}
We propose a novel framework for pre-training a generative model by leveraging a frozen Multimodal Large Language Model (MLLM) with an image-only dataset (see~\secref{sec:image_only_training}), entirely eschewing the need for paired text.
Our approach hinges on two key contributions.
First, to adapt the MLLM's representations for the generative task without costly fine-tuning, we introduce the \emph{\textbf{Residual Query Adapter}} (see~\secref{sec:residual_query_adapter}), a lightweight, parameter-efficient module that refines the visual condition.
Second, to prevent the self-conditioning from collapsing to a trivial identity mapping, we employ a \emph{\textbf{Masked Image Modeling}} strategy (see~\secref{sec:masked_image_modeling}). This transforms training into a sparse-to-dense reconstruction task, compelling the model to learn a robust and compositional visual prior.

\subsection{Preliminaries on Diffusion Models}
\label{sec:preliminaries}
Diffusion-based generative models transform a simple prior distribution, e.g., a standard Gaussian $\cN(\0, \mI)$, into a complex data distribution by learning to reverse a predefined noise-corruption process. In this paper, we focus on flow matching (FM) models~\citep{lipman2022flow}, which have demonstrated strong performance in image generation~\citep{xie2024sana,sun2025unified}.

Flow matching models define a deterministic path from a data point $\xx$ to a noise vector $\zz \sim \cN(\0, \mI)$ via the interpolation $\xx_t = (1-t)\cdot\xx + t\cdot\zz$ for $t \in [0,1]$. A neural network $\mmF_{\mtheta}(\xx_t, t, \cc)$ is then trained to learn the constant-velocity vector field $\zz - \xx$ of this path. Formally, given a conditioning signal $\cc$, the objective is:
$\mathcal{L}(\mtheta) = \EEb{\xx, \zz, \cc, t}{\norm{ \mmF_{\mtheta}(\xx_t, t, \cc) - (\zz - \xx) }_2^2 }$.

For generation, one starts with a sample from the prior, $\xx_1 \sim \cN(\0, \mI)$, and integrates the learned vector field backward in time from $t=1$ to $t=0$. This is achieved by solving the probability flow ordinary differential equation (PF-ODE)~\citep{song2020score}:
$
    \frac{\dm \xx_t}{\dm t} = \mmF_{\mtheta}(\xx_t, t, \cc).
$
The solution at $t=0$ yields the final generated sample $\xx_0$.

\subsection{Image-Only Pre-training via Self-Conditioning}
\label{sec:image_only_training}
We hypothesize that explicit text is merely one possible modality for conveying the high-level semantic information necessary to guide image synthesis. The rich semantic content inherent in an image can itself serve as a sufficient conditioning signal. This principle allows us to design a training paradigm that relies exclusively on an unlabeled image corpus.

Our framework utilizes a pre-trained and frozen MLLM, which we denote as $\mg$. This MLLM includes a Vision Transformer (ViT) encoder, $\mv$, for processing visual inputs.
To generate an image $\mathbf{x}$, we first derive a conditioning signal directly from $\mathbf{x}$.

\paragraph{Forming the self-conditioning signal.}
Inspired by instruction-following models, we construct the initial condition by combining a generic, fixed textual prompt with the visual features of the image.
Let $\mathbf{c}_{\mathrm{aux}} \in \mathbb{R}^{T \times D}$ be the token embeddings for an auxiliary prompt, such as ``\texttt{Generate an image that is identical to the reference image:}''.
The ViT encoder $\mv$ processes the image $\mathbf{x}$ into a sequence of patch embeddings, $\mathbf{c}_{\mathrm{img}} = \mv(\mathbf{x}) \in \mathbb{R}^{P^2 \times D}$, where $P^2$ is the number of patches and $D$ is the embedding dimension.

The complete conditioning sequence $\mathbf{c}$ is formed by concatenating these two components:
$
    \mathbf{c} = \mathrm{concat}(\mathbf{c}_{\mathrm{aux}}, \mathbf{c}_{\mathrm{img}}) \in \mathbb{R}^{(T+P^2) \times D}
$.
This sequence is then processed by the frozen MLLM $\mg$ to produce the final latent condition $\hh= \mg(\mathbf{c})$, which is used to guide the diffusion model $\mmF_{\mtheta}$.

\subsection{Residual Query Adapter}
\label{sec:residual_query_adapter}
Directly using the output of a frozen MLLM, $\mg(\mathbf{c})$, as a condition for the diffusion model yields suboptimal performance (see ``Raw'' in~\figref{fig:setting_performance}).
We attribute this to a domain mismatch: representations from an MLLM pre-trained for understanding-based tasks are not inherently optimized for the nuanced control required by a generative process.

While fine-tuning the entire MLLM ($\mg$) could in principle align its representations, this approach is fraught with two major challenges:
\begin{enumerate}[label=(\alph*), nosep, leftmargin=16pt]
    \item the immense computational cost associated with billions of parameters, where e.g.\ the MLLM in MetaQuery-XL has 7B parameters, versus 0.6B for the diffusion model~\citep{pan2025transfer}. \looseness=-1
    \item the risk of catastrophic forgetting, where the powerful, pre-trained capabilities of the MLLM are degraded when fine-tuned on an image-only reconstruction task.
\end{enumerate}
To circumvent these issues, we introduce the \textbf{Residual Query Adapter (RQA)}, denoted $\mq_{\mtheta}$.
The RQA is a lightweight (with only 29M parameters), trainable adapter module designed to preprocess the conditioning signal $\mathbf{c}$ before it enters the MLLM.
Specifically, the RQA uses cross-attention~\citep{vaswani2017attention} with 256 learned query tokens that learns a task-specific transformation.
It generates a ``residual query'' that is appended to the original conditioning sequence:
$
    \mathbf{c} \gets \text{concat}(\mathbf{c}, \mq_{\mtheta}(\mathbf{c}))
$.
The MLLM then processes this refined sequence, $\mathbf{h} = \mg(\mathbf{c})$.
The RQA acts as a learnable ``prompt'', guiding the frozen MLLM to extract features that are more salient for the downstream generative task without modifying any of the MLLM's original weights.

This parameter-efficient approach effectively adapts the MLLM for generation at a fraction of the computational cost.
The efficacy of the RQA is empirically validated in~\figref{fig:setting_performance} and~\secref{sec:ablation}.

\begin{algorithm}[t]
    \caption{Image-Only Pre-training for UMM Generation}
    \label{alg:training_framework}
    \begin{algorithmic}[1]
        \REQUIRE Image dataset $D$; frozen pre-trained MLLM $\mg$; frozen ViT encoder $\mv$; auxiliary prompt embeddings $\mathbf{c}_{\text{aux}}$; mask ratio $r$.
        \REQUIRE Randomly initialized diffusion network $\mmF_{\mtheta}$ and residual query adapter $\mq_{\mtheta}$.
        \REPEAT
        \STATE Sample image $\mathbf{x} \sim D$, noise $\mathbf{z} \sim \mathcal{N}(\mathbf{0}, \mathbf{I})$, time $t \sim \mathcal{U}(0, 1)$.
        \STATE Compute noised image: $\mathbf{x}_t = (1-t)\cdot\mathbf{x} + t\cdot\mathbf{z}$.
        \STATE Extract image patch embeddings: $\mathbf{c}_{\text{img}} = \mv(\mathbf{x})$.
        \STATE Generate random mask $\mathbf{M}$ with masking ratio $r$ and apply it: $\mathbf{c}_{\text{img}} \gets \mathbf{c}_{\text{img}} \odot \mathbf{M}$.
        \STATE Form the initial condition: $\mathbf{c} = \text{concat}(\mathbf{c}_{\text{aux}}, \mathbf{c}_{\text{img}})$.
        \STATE Refine condition with residual query adapter: $\mathbf{c} \gets \text{concat}(\mathbf{c}, \mq_{\mtheta}(\mathbf{c}))$.
        \STATE Compute latent condition from frozen MLLM: $\mathbf{h} = \mg(\mathbf{c})$.
        \STATE Compute loss: $\mathcal{L}(\mtheta) = \norm{ \mmF_{\mtheta}(\mathbf{x}_t, t, \mathbf{h}) - (\mathbf{z} - \mathbf{x}) }_2^2$.
        \STATE Update trainable parameters $\mtheta$ using gradients from $\mathcal{L}(\mtheta)$.
        \UNTIL{convergence}
    \end{algorithmic}
\end{algorithm}

\subsection{Masked Image Modeling}
\label{sec:masked_image_modeling}

A key feature of text-to-image training is the inherent sparsity of supervision: a short textual description provides only a high-level, incomplete specification of the corresponding image~\citep{xie2024sana,labs2025kontext}.
This forces the model to learn a compositional understanding of scenes and objects to fill in the missing details.
In contrast, our self-conditioning approach provides a dense, complete representation of the target image, which can encourage the model to learn a trivial identity mapping rather than a meaningful generative prior.

To emulate the benefits of sparse supervision, we introduce a \textbf{Masked Image Modeling} strategy inspired by masked autoencoders~\citep{he2022masked}.
During training, we randomly mask a fraction of the image patch tokens $\mathbf{c}_{\text{img}}$ with a masking ratio $r \in [0, 1]$. This is implemented by element-wise multiplication with a binary mask $\mathbf{M} \in \{0, 1\}^{P^2 \times D}$, where entries are drawn from a Bernoulli distribution with parameter $(1 - r)$:
$
    \mathbf{c}_{\text{img}} \gets \mathbf{c}_{\text{img}} \odot \mathbf{M}
$.
This simple yet effective technique transforms the training objective from dense reconstruction to a more challenging sparse-to-dense task.
The model is forced to infer the content of the masked patches from the visible ones, promoting the learning of robust, context-aware visual representations.
As shown in our experiments (see~\figref{fig:setting_performance} and \secref{sec:ablation}), this significantly improves generation quality.
Our complete training procedure is detailed in~\algref{alg:training_framework} and \figref{fig:framework}.

\begin{table*}[t]
    \centering
    \caption{\small{
            \textbf{Quantitative comparison on text-to-image generation benchmarks.}
            The ($\uparrow$) symbol indicates that higher scores are better.
            $^\dag$Results obtained using rewritten prompts from the original GenEval benchmark.
            $^*$Indicates the model was trained on an additional 30M proprietary image-text pairs.
        }
    }
    \vspace{-0.5em}
    \label{tab:geneval_benchmark}
    \resizebox{\textwidth}{!}{%
        \begin{tabular}{lccccccccc}
            \toprule
            \multirow{2}{*}{\textbf{METHOD}}        & \multicolumn{7}{c}{\textbf{GenEval}} & \multirow{2}{*}{\textbf{DPGBench} ($\uparrow$)} & \multirow{2}{*}{\textbf{WISE} ($\uparrow$)}                                                                                                              \\
                                                    & \textbf{Single Obj.}                 & \textbf{Two Obj.}                               & \textbf{Counting}                           & \textbf{Colors} & \textbf{Position} & \textbf{Color Attri.} & \textbf{Overall} ($\uparrow$) &       &      \\
            \midrule
            \multicolumn{10}{c}{\textbf{Gen. Only}}                                                                                                                                                                                                                                                     \\
            \midrule
            SDv1.5~\citep{rombach2022high}          & 0.97                                 & 0.38                                            & 0.35                                        & 0.76            & 0.04              & 0.06                  & 0.43                          & 63.18 & 0.32 \\
            SDv2.1~\citep{rombach2022high}          & 0.98                                 & 0.51                                            & 0.44                                        & 0.85            & 0.07              & 0.17                  & 0.50                          & -     & 0.32 \\
            SD3-Medium~\citep{esser2024scaling}     & 0.99                                 & 0.94                                            & 0.72                                        & 0.89            & 0.33              & 0.60                  & 0.74                          & 84.08 & 0.42 \\
            SDXL~\citep{podell2024sdxl}             & 0.98                                 & 0.74                                            & 0.39                                        & 0.85            & 0.15              & 0.23                  & 0.55                          & 74.65 & 0.43 \\
            PixArt-$\alpha$~\citep{chen2024pixart}  & 0.98                                 & 0.50                                            & 0.44                                        & 0.80            & 0.08              & 0.07                  & 0.48                          & 71.11 & 0.47 \\
            DALL-E 2~\citep{ramesh2022hierarchical} & 0.94                                 & 0.66                                            & 0.49                                        & 0.77            & 0.10              & 0.19                  & 0.52                          & -     & -    \\
            DALL-E 3~\citep{dalle3}                 & 0.96                                 & 0.87                                            & 0.47                                        & 0.83            & 0.43              & 0.45                  & 0.67                          & 83.50 & -    \\
            Lumos-T2I~\citep{ma2025learning}        & 0.99                                 & 0.64                                            & 0.52                                        & 0.84            & 0.15              & 0.30                  & 0.57                          & 79.90 & -    \\
            \midrule
            \multicolumn{10}{c}{\textbf{Unified Models}}                                                                                                                                                                                                                                                \\
            \midrule
            Chameleon~\citep{team2024chameleon}     & -                                    & -                                               & -                                           & -               & -                 & -                     & 0.39                          & -     & -    \\
            Show-o~\citep{xie2025showo}             & 0.98                                 & 0.80                                            & 0.66                                        & 0.84            & 0.31              & 0.50                  & 0.68                          & -     & 0.35 \\
            Show-o2-7B~\citep{xie2025showo2}        & 1.00                                 & 0.87                                            & 0.58                                        & 0.92            & 0.52              & 0.62                  & 0.76$\dag$                    & 86.14 & 0.39 \\
            Janus~\citep{wu2024janus}               & 0.97                                 & 0.68                                            & 0.30                                        & 0.84            & 0.46              & 0.42                  & 0.61                          & 79.68 & 0.23 \\
            JanusFlow~\citep{ma2024janusflow}       & 0.97                                 & 0.59                                            & 0.45                                        & 0.83            & 0.53              & 0.42                  & 0.63                          & 80.09 & -    \\
            Janus-Pro-1B~\citep{chen2025januspro}   & 0.98                                 & 0.82                                            & 0.51                                        & 0.89            & 0.65              & 0.56                  & 0.73                          & 82.63 & 0.26 \\
            Janus-Pro-7B~\citep{chen2025januspro}   & 0.99                                 & 0.89                                            & 0.59                                        & 0.90            & 0.79              & 0.66                  & 0.80                          & 84.19 & 0.35 \\
            MetaQuery-B~\citep{pan2025transfer}     & -                                    & -                                               & -                                           & -               & -                 & -                     & 0.74$\dag$                    & 80.04 & 0.46 \\
            MetaQuery-L~\citep{pan2025transfer}     & -                                    & -                                               & -                                           & -               & -                 & -                     & 0.78$\dag$                    & 81.10 & 0.55 \\
            MetaQuery-XL~\citep{pan2025transfer}    & -                                    & -                                               & -                                           & -               & -                 & -                     & 0.80$\dag$                    & 82.05 & 0.55 \\
            BLIP3-o-4B~\citep{chen2025blipo}        & -                                    & -                                               & -                                           & -               & -                 & -                     & 0.81                          & 79.36 & 0.50 \\
            BLIP3-o-8B*~\citep{chen2025blipo}       & -                                    & -                                               & -                                           & -               & -                 & -                     & 0.84                          & 81.60 & 0.62 \\
            BAGEL-7B~\citep{deng2025emerging}       & 0.98                                 & 0.95                                            & 0.84                                        & 0.95            & 0.78              & 0.77                  & 0.88$\dag$                    & -     & 0.52 \\

            \midrule
            \multicolumn{10}{c}{\textbf{Ours}}                                                                                                                                                                                                                                                          \\
            \midrule
            \method-B 512                           & 0.99                                 & 0.92                                            & 0.83                                        & 0.94            & 0.91              & 0.75                  & 0.89                          & 82.95 & 0.55 \\
            \method-B 1024                          & 0.99                                 & 0.91                                            & 0.75                                        & 0.93            & 0.88              & 0.75                  & 0.87                          & 80.71 & 0.50 \\
            \method-L 512                           & 0.99                                 & 0.91                                            & 0.82                                        & 0.94            & 0.85              & 0.72                  & 0.87                          & 76.09 & 0.53 \\
            \method-L 1024                          & 1.00                                 & 0.91                                            & 0.71                                        & 0.92            & 0.78              & 0.78                  & 0.85                          & 72.26 & 0.48 \\
            \bottomrule
        \end{tabular}%
    }
    \vspace{-0.5em}
\end{table*}

\section{Experiment}
\label{sec:exp}
We conduct comprehensive experiments to validate the efficacy of our proposed framework, \method. Our evaluation is designed to systematically assess its performance in text-to-image generation, analyze the impact of different training data compositions, and ablate its core architectural components.
\looseness=-1

\subsection{Experimental Setting} \label{sec:expset}
\paragraph{Datasets.}
Our pre-training corpus comprises the Megalith-10M~\citep{matsubara2024megalith10m} and text-to-image-2M~\citep{text2image2m_2024} datasets. For the fine-tuning stage, we leverage a curated collection of high-quality, instruction-following datasets, namely BLIP3-o-60K~\citep{chen2025blipo}, Echo-4o-Image~\citep{ye2025echoo}, and ShareGPT-4o-Image~\citep{chen2025sharegptoimage}.
All images undergo a standardized preprocessing pipeline: we apply a central crop and resize them to a resolution of either $512 \times 512$ or $1024 \times 1024$. 

\paragraph{Neural network architectures.}
The core of our model adopts the Multi-Modal Diffusion Transformer (MM-DiT) architecture \citep{esser2024scaling}, as implemented in FLUX \citep{labs2025flux1kontextflowmatching}. This design employs independent attention mechanisms for image and text modalities to facilitate robust cross-modal fusion. To investigate scaling properties, we instantiate three variants: \method-B ($1.6\mathrm{B}$ parameters), \method-L ($2.7\mathrm{B}$ parameters), and \method-XL, with the latter following the $6\mathrm{B}$ parameter Z-Image framework \citep{cai2025z}. For the auxiliary MLLM component, a frozen InternVL3-2B \citep{zhu2025internvl3} is employed as a feature extractor, offering high-quality representations with a minimal computational footprint.

\looseness=-1
\paragraph{Implementation and evaluation.}
We implement our framework in PyTorch~\citep{paszke2019pytorch} and utilize the AdamW optimizer~\citep{loshchilov2017decoupled} for training of \method-B and \method-L and the Muon optimizer~\citep{jordan2024muon} for \method-XL.
Adhering to established practices in generative modeling~\citep{yu2024representation,ma2024sit}, we maintain an exponential moving average (EMA) of the model weights with a decay rate of $0.999$.
All reported results are derived from the EMA model weights to ensure stability and improved performance.
For evaluation, we follow standard protocols established in prior works~\citep{pan2025transfer,chen2025januspro,esser2024scaling}.
To assess generative quality and text-image alignment, we employ a suite of comprehensive benchmarks: GenEval~\citep{ghosh2023geneval}, DPG-Bench~\citep{hu2024ella}, and WISE~\citep{niu2025wise}.
The image editing capabilities of our model are specifically evaluated using the ImgEdit-Bench~\citep{ye2025imgedit}.
Further details regarding hyperparameters and the training infrastructure are available in~\appref{app:detailed_exp}.

\subsection{Performance on Text-to-Image Generation}
\label{sec:text2image}

\begin{table*}[t]
    \centering
    \caption{\small{\textbf{Evaluating different fine-tuning strategies on various open-source UMMs.}
            The notation A \rp\ B denotes applying fine-tuning method B to a pre-trained model A. The symbols \dg{}/\up{} indicate the performance change relative to the baseline pre-trained model.}}
    \vspace{-0.5em}
    \label{tab:umm_finetuning}
    \resizebox{\textwidth}{!}{%
        \begin{tabular}{lcc|cccccccc}
            \toprule
            \multirow{2}{*}{\textbf{METHOD}}                  & \multirow{2}{*}{\textbf{Res.}} & \multirow{2}{*}{\textbf{NFE}} & \multicolumn{7}{c}{\textbf{GenEval}} & \multirow{2}{*}{\textbf{WISE} ($\uparrow$)}                                                                                                                                             \\
                                                              &                                &                               & \textbf{Single Obj.}                 & \textbf{Two Obj.}                           & \textbf{Counting}  & \textbf{Colors}    & \textbf{Position}  & \textbf{Color Attri.} & \textbf{Overall} ($\uparrow$) &                    \\
            \midrule
            OpenUni-L~\citep{wu2025openuni}                   & 512                            & 20$\times$2                   & 0.99                                 & 0.91                                        & 0.77               & 0.90               & 0.75               & 0.76                  & 0.85                          & 0.52               \\
            \fd{\quad \rp Image finetuning}                   & \fd{512}                       & \fd{20$\times$2}              & \fd{1.00\up{0.01}}                   & \fd{0.98\up{0.07}}                          & \fd{0.22\dg{0.55}} & \fd{0.91\up{0.01}} & \fd{0.60\dg{0.15}} & \fd{0.77\up{0.01}}    & \fd{0.74\dg{0.11}}            & \fd{0.49\dg{0.03}} \\
            \quad \rp Pair finetuning                         & 512                            & 20$\times$2                   & 0.99                                 & 0.94\up{0.03}                               & 0.82\up{0.05}      & 0.91\up{0.01}      & 0.85\up{0.10}      & 0.76                  & 0.88\up{0.03}                 & 0.62\up{0.10}      \\
            \quad \rp Mix finetuning                          & 512                            & 20$\times$2                   & 0.99                                 & 0.91                                        & 0.78\up{0.01}      & 0.93\up{0.03}      & 0.87\up{0.12}      & 0.78\up{0.02}         & 0.88\up{0.03}                 & 0.59\up{0.07}      \\
            \midrule
            Qwen-Image~\citep{wu2025qwenimagetechnicalreport} & 512                            & 50$\times$2                   & 0.99                                 & 0.91                                        & 0.87               & 0.88               & 0.73               & 0.74                  & 0.85                          & -                  \\
            \fd{\quad \rp Image finetuning}                   & \fd{512}                       & \fd{50$\times$2}              & \fd{0.55\dg{0.44}}                   & \fd{0.51\dg{0.40}}                          & \fd{0.38\dg{0.49}} & \fd{0.43\dg{0.45}} & \fd{0.30\dg{0.43}} & \fd{0.37\dg{0.37}}    & \fd{0.42\dg{0.43}}            & \fd{0.41}          \\
            \quad \rp Pair finetuning                         & 512                            & 50$\times$2                   & 1.00\up{0.01}                        & 0.93\up{0.02}                               & 0.88\up{0.01}      & 0.91\up{0.03}      & 0.82\up{0.09}      & 0.75\up{0.01}         & 0.88\up{0.03}                 & 0.63               \\
            \quad \rp Mix finetuning                          & 512                            & 50$\times$2                   & 1.00\up{0.01}                        & 0.92\up{0.01}                               & 0.87               & 0.91\up{0.03}      & 0.82\up{0.09}      & 0.79\up{0.05}         & 0.89\up{0.04}                 & 0.63               \\
            \midrule
            Qwen-Image~\citep{wu2025qwenimagetechnicalreport} & 1024                           & 50$\times$2                   & 0.99                                 & 0.93                                        & 0.88               & 0.90               & 0.77               & 0.74                  & 0.87                          & 0.62               \\
            \fd{\quad \rp Image finetuning}                   & \fd{1024}                      & \fd{50$\times$2}              & \fd{0.54\dg{0.45}}                   & \fd{0.61\dg{0.32}}                          & \fd{0.47\dg{0.41}} & \fd{0.47\dg{0.43}} & \fd{0.28\dg{0.49}} & \fd{0.47\dg{0.27}}    & \fd{0.47\dg{0.40}}            & \fd{0.35\dg{0.27}} \\
            \quad \rp Pair finetuning                         & 1024                           & 50$\times$2                   & 1.00\up{0.01}                        & 0.93\up{0.01}                               & 0.88               & 0.91\up{0.01}      & 0.82\up{0.05}      & 0.75\up{0.01}         & 0.88\up{0.01}                 & 0.63\up{0.01}      \\
            \quad \rp Mix finetuning                          & 1024                           & 50$\times$2                   & 0.99                                 & 0.92\dg{0.01}                               & 0.90\up{0.02}      & 0.91\up{0.01}      & 0.81\up{0.04}      & 0.80\up{0.06}         & 0.89\up{0.02}                 & 0.63\up{0.01}      \\
            \bottomrule
        \end{tabular}%
    }
    \vspace{-1em}
\end{table*}

\begin{table*}[t!]
    \centering
    \caption{\small{\textbf{Image editing benchmark results.} Methods highlighted in \colorbox{red!10}{red} are trained on specific editing datasets. Our \method, highlighted in \colorbox{blue!10}{blue}, is evaluated in a \colorbox{blue!10}{training-free} setting without any training on editing data.}}
    \vspace{-0.5em}
    \label{tab:img_edit_result}
    \resizebox{\textwidth}{!}{%
        \begin{tabular}{l|cccccccccc}
            \toprule
            \multirow{2}{*}{\textbf{METHOD}}                                     & \multicolumn{10}{c}{\textbf{ImgEdit-Bench}}                                                                                                                                                                                      \\

                                                                                 & \textbf{Add}                                & \textbf{Adjust} & \textbf{Extract} & \textbf{Replace} & \textbf{Remove} & \textbf{Background} & \textbf{Style} & \textbf{Hybrid} & \textbf{Action} & \textbf{Overall} ($\uparrow$) \\
            \midrule
            \multicolumn{11}{c}{\textbf{Trained with editing data}}                                                                                                                                                                                                                                                 \\
            \midrule
            \rowcolor{red!10} MagicBrush~\citep{zhang2023magicbrush}             & 2.84                                        & 1.58            & 1.51             & 1.97             & 1.58            & 1.75                & 2.38           & 1.62            & 1.22            & 1.90                          \\
            \rowcolor{red!10} Instruct-Pix2Pix \citep{brooks2023instructpix2pix} & 2.45                                        & 1.83            & 1.44             & 2.01             & 1.50            & 1.44                & 3.55           & 1.20            & 1.46            & 1.88                          \\
            \rowcolor{red!10} AnyEdit~\citep{yu2025anyedit}                      & 3.18                                        & 2.95            & 1.88             & 2.47             & 2.23            & 2.24                & 2.85           & 1.56            & 2.65            & 2.45                          \\
            \rowcolor{red!10} UltraEdit~\citep{zhao2024ultraedit}                & 3.44                                        & 2.81            & 2.13             & 2.96             & 1.45            & 2.83                & 3.76           & 1.91            & 2.98            & 2.70                          \\
            \rowcolor{red!10} OmniGen~\citep{xiao2025omnigen}                    & 3.47                                        & 3.04            & 1.71             & 2.94             & 2.43            & 3.21                & 4.19           & 2.24            & 3.38            & 2.96                          \\
            \rowcolor{red!10} ICEdit~\citep{zhang2025context}                    & 3.58                                        & 3.39            & 1.73             & 3.15             & 2.93            & 3.08                & 3.84           & 2.04            & 3.68            & 3.05                          \\
            \rowcolor{red!10} Step1X-Edit~\citep{liu2025step1x}                  & 3.88                                        & 3.14            & 1.76             & 3.40             & 2.41            & 3.16                & 4.63           & 2.64            & 2.52            & 3.06                          \\
            \rowcolor{red!10} BAGEL~\citep{deng2025emerging}                     & 3.56                                        & 3.31            & 1.70             & 3.3              & 2.62            & 3.24                & 4.49           & 2.38            & 4.17            & 3.20                          \\
            \midrule
            \multicolumn{11}{c}{\textbf{Ours (zero-shot)}}                                                                                                                                                                                                                                                          \\
            \midrule
            \rowcolor{blue!10} \method-B (text-image pair pre-trained)           & 3.18                                        & 2.17            & 1.92             & 2.70             & 1.17            & 3.36                & 4.39           & 1.49            & 3.14            & 2.61                          \\
            \rowcolor{blue!10} \method-B (image-only pre-trained)                & 3.84                                        & 2.37            & 2.12             & 2.60             & 1.30            & 3.14                & 4.41           & 1.80            & 3.78            & 2.82                          \\
            \bottomrule
        \end{tabular}%
    }
    \vspace{-0.5em}
\end{table*}

We benchmark {\method} against SOTA models in~\tabref{tab:geneval_benchmark}.
Our base model, {\method-B (512px)} built on a 1.6B generative backbone, achieves a new SOTA score of $0.89$ on GenEval. Notably, this performance surpasses strong baselines like BAGEL ($0.88$) and BLIP3-o-8B*($0.84$, trained with an extra $30\mathrm{M}$ proprietary image-text pairs), despite {\method} being trained exclusively on public datasets and with remarkable efficiency ($1050$ H800 GPU hours). Furthermore, {\method-B} attains a competitive score of $0.55$ on the WISE benchmark, demonstrating that our approach effectively preserves world knowledge without degradation.
Qualitative results in~\figref{fig:main_image} showcase our model's strong compositional abilities.
\paragraph{Analysis of model scaling.}
The lower performance of our larger {\method-L} model is an artifact of constrained training resources; it was trained for half the epochs of {\method-B}.
When controlling for training duration ($5$ epochs), {\method-L} outperforms {\method-B} ($0.87$ vs.\ $0.86$ on GenEval), confirming a positive scaling trend and suggesting potential for further gains with continued training.

\begin{figure}[t!]
    \centering
    \includegraphics[width=\linewidth]{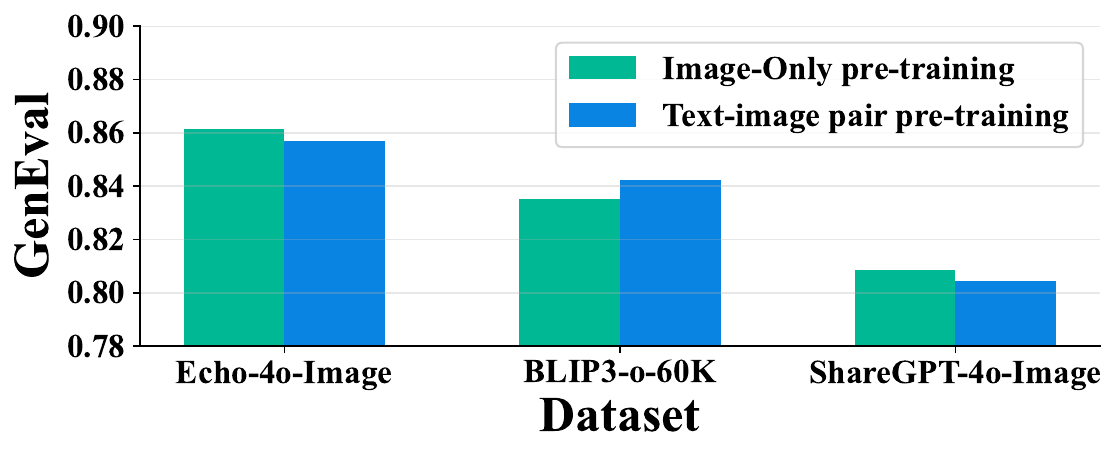}
    \vspace{-2.0em}
    \caption{\small{
            \textbf{Analysis of different data paradigms.}
            Fine-tuning performance comparison of models pre-trained on different data compositions (image-only, text-image pair) across distinct datasets.
        }}
    \label{fig:diff_recipe}
    \vspace{-1em}
\end{figure}

\begin{figure*}[t!]
    \centering
    \begin{subfigure}[b]{0.33\textwidth}
        \centering
        \includegraphics[width=\linewidth]{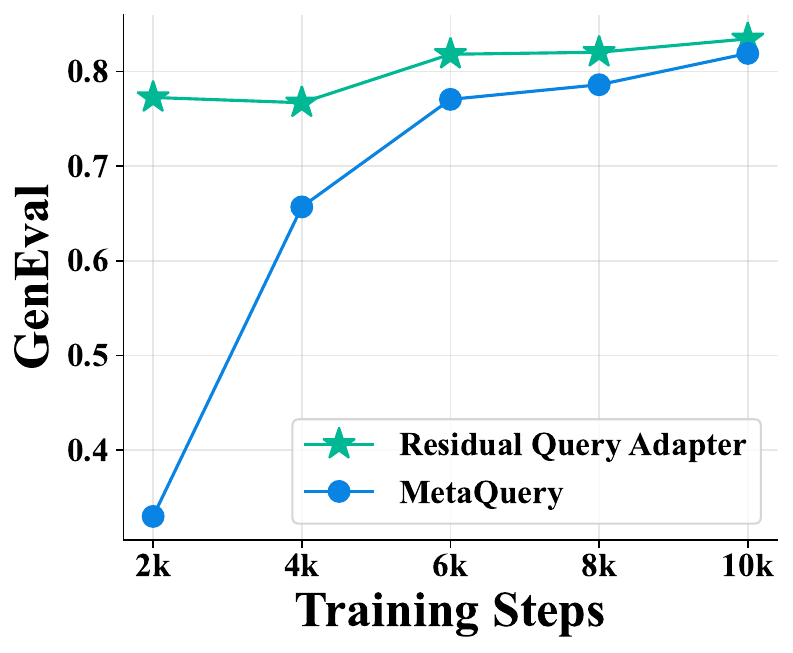}
        \caption{\textbf{Residual query adapter.}}
        \label{fig:diff_cross_attention}
    \end{subfigure}
    \hfill
    \begin{subfigure}[b]{0.33\textwidth}
        \centering
        \includegraphics[width=\linewidth]{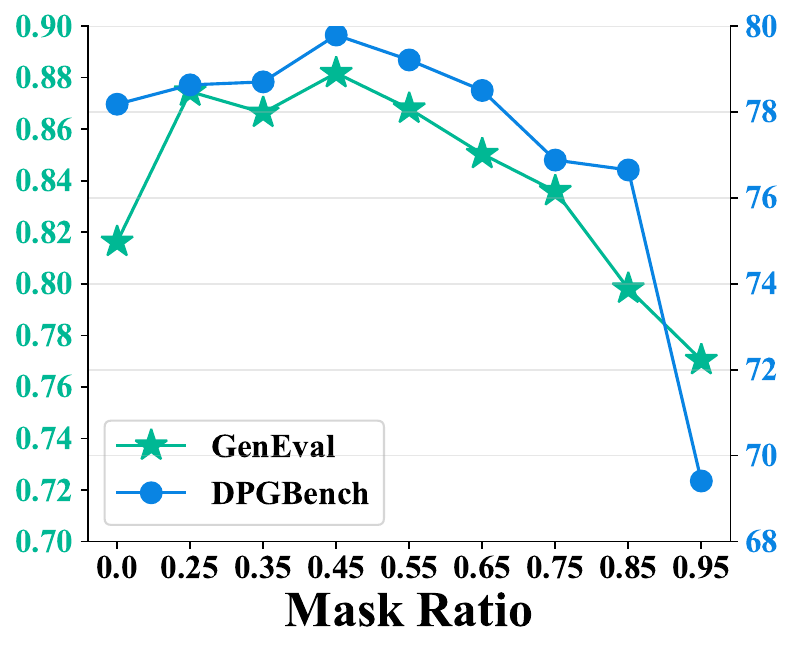}
        \caption{\textbf{Various mask ratio.}}
        \label{fig:diff_mask}
    \end{subfigure}
    \hfill
    \begin{subfigure}[b]{0.33\textwidth}
        \centering
        \includegraphics[width=\linewidth]{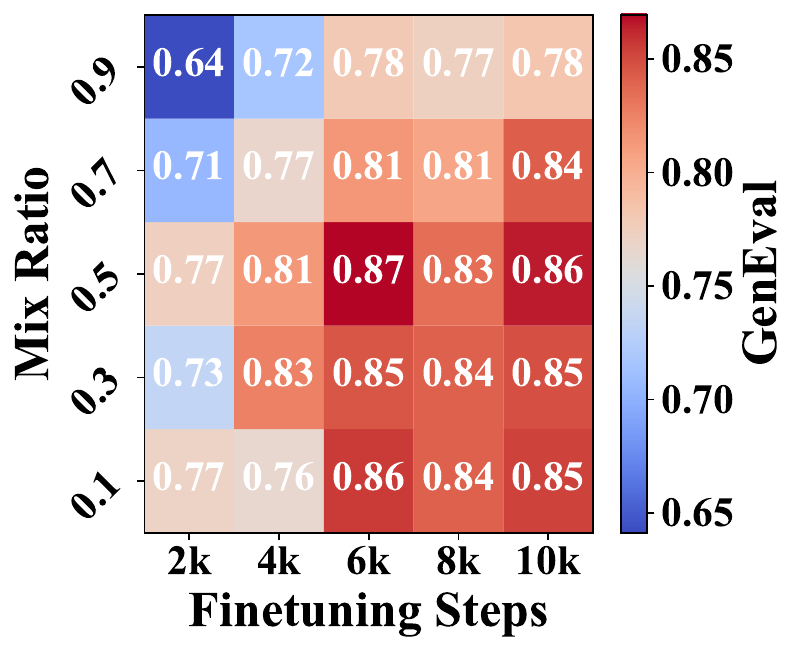}
        \caption{\textbf{Various mix ratio of data.}}
        \label{fig:diff_mix_ratio}
    \end{subfigure}
    \hfill
    \vspace{-0.5em}
    \caption{\small{
            \textbf{Ablation studies of key components in \method.}
            These experiments analyze the impact of our primary design choices: (a) the residual query adapter, (b) the mask ratio for sparse reconstruction, and (c) the data mixture ratio during fine-tuning.
        }}
    \vspace{-0.5em}
\end{figure*}

\subsection{Impact of Pre-training and Fine-tuning Data}
\label{sec:different_paradigms}

We investigate the impact of data composition during the pre-training and fine-tuning stages. We define three distinct data types: (a) image-only, (b) text-image pairs, and (c) a mixture of both.
This section presents a systematic ablation study on the six possible combinations of these data types across the two stages, focusing on their efficacy for text-to-image generation.
\looseness=-1
\paragraph{The role of pre-training data.}
We first compare models pre-trained on image-only data versus those pre-trained on text-image pairs. As illustrated in~\figref{fig:diff_recipe} and~\figref{fig:6_paradigm_result}, the image-only pre-trained model consistently achieves superior or comparable performance to its text-image pair counterpart, irrespective of the fine-tuning data composition.
\paragraph{The role of fine-tuning data.}
Next, we analyze the effect of the fine-tuning data composition. Beyond using image-only or text-image pair data exclusively, we explore a mixed-data strategy. Remarkably,~\figref{fig:6_paradigm_result} reveals that for models pre-trained under \textbf{both} paradigms, fine-tuning with the mixed data yields the highest performance on GenEval. Conversely, fine-tuning with image-only data consistently results in the lowest scores.

\looseness=-1
\paragraph{Generalization to open-source UMMs.}
To validate the generalizability of our findings, we apply our fine-tuning strategies to prominent open-source UMMs: OpenUni-L-3.6B~\citep{wu2025openuni} and Qwen-Image-20B~\citep{wu2025qwenimagetechnicalreport}.
For the larger Qwen-Image model, we employ LoRA~\citep{hu2022lora} (with $r=64$ and $\alpha=64$) for computational efficiency.
The results, summarized in~\tabref{tab:umm_finetuning}, corroborate our primary conclusion: the mixed-data fine-tuning approach consistently outperforms the other strategies on GenEval.
For instance, it improves the GenEval score of OpenUni-L from a baseline of $0.85$ to $0.88$. Even for the powerful Qwen-Image model, this strategy yields notable gains, increasing scores from $0.85$ to $0.89$ (512px) and $0.87$ to $0.89$ (1024px).

Beyond generation quality, we evaluate world knowledge and reasoning using the WISE benchmark. As shown in the final column of~\tabref{tab:umm_finetuning}, both text-image pair and mixed-data fine-tuning provide a substantial performance uplift for OpenUni-L (up to $0.10$) and a modest improvement for Qwen-Image ($0.01$). In contrast, fine-tuning with image-only data proves detrimental across nearly all scenarios, significantly impairing the models' prompt-following ability—an effect particularly pronounced in larger models (see~\appref{app:umm_finetune_result} for a detailed analysis).
\paragraph{Emergent image editing capabilities.}
A surprising and significant finding is the emergence of strong image editing capabilities.~\tabref{tab:img_edit_result} demonstrates that our model, when pre-trained on image-only data, achieves competitive performance on the ImgEdit-Bench benchmark. Crucially, this is accomplished in a \textit{zero-shot setting}, without any fine-tuning on task-specific editing data. This training-free approach not only surpasses the performance of the same model pre-trained on text-image pairs but also outperforms several strong baselines like UltraEdit~\citep{zhao2024ultraedit} that are explicitly trained on editing datasets.

\subsection{Ablation Studies on Key Components of \method}
\label{sec:ablation}

Unless specified otherwise, all experiments in this section are conducted using the \method-XL model pre-trained exclusively on image-only data.
\paragraph{Efficacy of the residual query adapter.}
To further validate the efficacy of our proposed residual query adapter, we compare it against a strong baseline, MetaQuery~\citep{pan2025transfer}, trained on identical data with the same 256 query tokens. The results, depicted in~\figref{fig:diff_cross_attention}, clearly demonstrate that our approach achieves a significantly faster convergence rate.
Notably, extending the fine-tuning of MetaQuery by an additional 8K steps only yields a score of $0.82$ on GenEval.
\paragraph{Impact of image token mask ratio.}
We investigate the impact of the mask ratio for image tokens, a key parameter in our sparse reconstruction objective. As shown in~\figref{fig:diff_mask}, performance improves as the ratio increases, peaking at an impressive 0.88 GenEval score and a DPGBench score of 79.79 with a mask ratio of 0.45. This result validates the effectiveness of our learning paradigm. However, an excessively high ratio (e.g., 0.95) leads to a sharp performance degradation (a drop to 0.77 and 69.41, respectively), likely due to significant information loss that impairs the training guidance for the generation process.
\paragraph{Influence of data mixture ratio.}
We examine the effect of varying the proportion of image-only data versus text-image pairs during the fine-tuning stage. A mix ratio of 1.0 corresponds to pure image-only data, while 0.0 signifies pure text-image pairs.~\figref{fig:diff_mix_ratio} reveals that performance initially increases with the mix ratio, reaching its optimum at 0.5. Furthermore, an optimal ratio of approximately 0.5 not only yields the best results but also demonstrates greater training stability, whereas lower ratios are prone to performance volatility in the later stages of fine-tuning.

\section{Conclusion}
We introduced \method, a novel and efficient framework for training UMM visual generation components using primarily \textbf{image-only data}, addressing the common paired-data bottleneck. Our two-stage approach---image-only pre-training followed by mixed-data fine-tuning---achieves SOTA performance with remarkable computational efficiency. Furthermore, we demonstrate that our \textbf{mixed-data fine-tuning} strategy is a generalizable technique that consistently enhances the performance of existing powerful UMMs. Detailed settings and results are in~\appref{app:detailed_exp}.

%% file: resources/appendix.tex
\onecolumn
{
    \hypersetup{linkcolor=black}
    \parskip=0em
    \renewcommand{\contentsname}{Contents}
    \tableofcontents
    \addtocontents{toc}{\protect\setcounter{tocdepth}{3}}
}

\newpage

\section{Utilization of Large Language Models (LLMs)}
In this study, Large Language Models (LLMs) are employed at the sentence level to assist in linguistic refinement. Their use was strictly confined to improving grammatical accuracy and overall readability of the manuscript. All research concepts, methodological designs, experimental processes, and analytical findings remain entirely original and have been solely contributed by the authors.

\section{Detailed Experimental Settings}
\label{app:detailed_exp}
This section elaborates on the experimental setup, including all relevant hyperparameter choices.

\subsection{Pre-training Settings}
\label{app:pretraining_settings}
The results presented in \tabref{tab:geneval_benchmark} are derived using the pre-training configurations outlined in \tabref{tab:pretraining_settings}. Due to computational resource constraints, Exponential Moving Average (EMA) decay was not applied during the training of \method-L and \method-XL. All models were pre-trained on the Megalith-10M~\citep{matsubara2024megalith10m} and text-to-image-2M~\citep{text2image2m_2024} datasets (except for \method-XL), comprising approximately 11 million images in total. Each image was resized so that its shorter edge was $512$ pixels while preserving the original aspect ratio, then a central crop was applied to obtain a $512 \times 512$ image. 
Notably, since neither dataset provides images at a resolution of $1024 \times 1024$, we did not deploy high-resolution pre-training.
\begin{table*}[h]
    \centering
    \caption{\small{\textbf{Pre-training settings.}}}
    \vspace{-0.5em}
    \label{tab:pretraining_settings}
    \begin{tabular}{l|cc|c}
        \toprule
        \textbf{METHOD}            & \textbf{\method-B}               & \textbf{\method-L}               & \textbf{\method-XL} \\
        \midrule
        \multicolumn{4}{c}{Optimization}                                                   \\
        \midrule
        Optimizer                  & \multicolumn{2}{c|}{AdamW}                             & Muon \\
        $\beta$                    & \multicolumn{2}{c|}{$(0.9,0.95)$}                      & $(0.9,0.95)$  \\
        Learning rate              & \multicolumn{2}{c|}{1e-4}                              & 1e-4 \\
        Max gradient norm          & \multicolumn{2}{c|}{1.0}                               & 1.0 \\
        Weight decay               & \multicolumn{2}{c|}{0.0}                               & 0.0 \\
        \midrule
        \multicolumn{4}{c}{Training Configuration}                                         \\
        \midrule
        Generative Model Size      & 1.6B                             & 2.7B               & 6B \\
        Training data type         & Image-only                       & Image-only         & Image-only \\
        EMA decay                  & 0.999                            & -                  & - \\
        Global batch size          & 1024                             & 512                & 4096 \\
        Image token mask ratio $r$ & 0.85                             & 0.85               & 0.45 \\

        \bottomrule
    \end{tabular}%
    \vspace{-1em}
\end{table*}

\subsection{Finetuning Settings}
\label{app:finetuning_settings}
We fine-tuned the two models (B\&L) at resolutions of 512 and 1024, respectively, using the pre-training settings specified in \tabref{tab:pretraining_settings}. The fine-tuning datasets include BLIP3o-60K~\citep{chen2025blipo}, Echo-4o-Image~\citep{ye2025echoo}, and ShareGPT-4o-Image~\citep{chen2025sharegptoimage}, collectively comprising approximately 210,000 high-resolution images (except for \method-XL). All images in these datasets are at $1024\times 1024$ resolution. For fine-tuning at both 512 and 1024 resolutions, we applied central cropping to resize images to the target resolution. 
\begin{table*}[h]
    \centering
    \caption{\small{\textbf{Finetuning settings.}}}
    \vspace{-0.5em}
    \label{tab:finetuning_settings}
    \begin{tabular}{l|cc|cc|c}
        \toprule
        \textbf{METHOD}            & \multicolumn{2}{c}{\textbf{\method-B}} & \multicolumn{2}{|c}{\textbf{\method-L}}               & \textbf{\method-XL} \\
        \textbf{Resolution}        & 512                                    & 1024                                    & 512  & 1024 & 512 \\
        \midrule
        \multicolumn{6}{c}{Optimization}                                                                                            \\
        \midrule
        Optimizer                  & \multicolumn{2}{c|}{AdamW}             & \multicolumn{2}{c|}{AdamW}                             & Muon \\
        $\beta$                    & \multicolumn{2}{c|}{$(0.9,0.95)$}      & \multicolumn{2}{c|}{$(0.9,0.95)$}                      & $(0.9,0.95)$  \\
        Learning rate              & \multicolumn{2}{c|}{1e-4}              & \multicolumn{2}{c|}{1e-4}                              & 1e-4 \\
        Max gradient norm          & \multicolumn{2}{c|}{1.0}               & \multicolumn{2}{c|}{1.0}                               & 1.0 \\
        Weight decay               & \multicolumn{2}{c|}{0.0}               & \multicolumn{2}{c|}{0.0}                               & 0.0 \\
        Generative Model Size      & 1.6B                                   & 1.6B                                    & 2.7B & 2.7B & 6B \\
        \midrule
        \multicolumn{6}{c}{Training Configuration}                                                                                  \\
        \midrule
        Training data type         & Mix                                    & Mix                                     & Mix  & Mix  & Mix \\
        EMA decay                  & 0.999                                  & 0.999                                   & -    & -    & - \\
        Global batch size          & 256                                    & 96                                      & 256  & 96   & 256 \\
        Image token mask ratio $r$ & 0.85                                   & 0.85                                    & 0.85 & 0.85 & 0.45 \\
        Mix ratio $\lambda$        & 0.5                                    & 0.5                                     & 0.5  & 0.5  & 0.5 \\
        \bottomrule
    \end{tabular}%
    \vspace{-1em}
\end{table*}

\subsection{UMM Finetuning Settings}
\label{app:umm_finetuning_settings}
The results presented in \tabref{tab:umm_finetuning} were obtained using the fine-tuning configurations specified in \tabref{tab:umm_finetuning_settings}. For OpenUni-L, we performed full fine-tuning on both the connector module and the generative model. In contrast, for Qwen-Image-20B, we applied Low-Rank Adaptation (LoRA)~\citep{hu2022lora} to fine-tune the model. Both models utilized a frozen understanding module. Additionally, due to computational constraints, Exponential Moving Average (EMA) decay was not implemented for Qwen-Image-20B.
\begin{table*}[h]
    \centering
    \caption{\small{\textbf{UMM finetuning settings.}}}
    \vspace{-0.5em}
    \label{tab:umm_finetuning_settings}
    \begin{tabular}{l|cc}
        \toprule
        \textbf{METHOD}            & OpenUni-L           & Qwen-Image-20B      \\
        \midrule
        \multicolumn{3}{c}{Optimization}                                       \\
        \midrule
        Optimizer                  & AdamW               & AdamW               \\
        $\beta$                    & (0.9,0.95)          & (0.9,0.95)          \\
        Learning rate              & 1e-4                & 1e-4                \\
        Max gradient norm          & 1.0                 & 1.0                 \\
        Weight decay               & 0.0                 & 0.0                 \\
        \midrule
        \multicolumn{3}{c}{Training Configuration}                             \\
        \midrule
        Training data type         & Mix/Image-only/Pair & Mix/Image-only/Pair \\
        EMA decay                  & 0.999               & -                   \\
        Global batch size          & 256                 & 48                  \\
        Epochs                     & 12                  & 5                   \\
        Image token mask ratio $r$ & 0.85                & 0.85                \\
        Mix ratio $\lambda$        & 0.5                 & 0.5                 \\
        \midrule
        \multicolumn{3}{c}{LoRA Configuration}                                 \\
        \midrule
        LoRA rank                  & -                   & 64                  \\
        LoRA alpha                 & -                   & 64                  \\
        LoRA dropout               & -                   & 0.0                 \\
        \bottomrule
    \end{tabular}%
    \vspace{-1em}
\end{table*}

\section{More Results}

\subsection{DPGBench Evaluation Results}
\label{app:dpg_benchmark}
The \tabref{tab:dpg_benchmark} shows the detailed results of the DPGBench evaluation shown in \tabref{tab:geneval_benchmark}.

\begin{table*}[h]
    \centering
    \caption{\small{\textbf{DPGBench evaluation results.} Here BLIP3-o-8B* donates the model that is trained with an 30 million proprietary data.}}
    \vspace{-0.5em}
    \label{tab:dpg_benchmark}
        \begin{tabular}{lcccccc}
            \toprule
            \textbf{METHOD}                        & \textbf{Global} & \textbf{Entity} & \textbf{Attribute} & \textbf{Relation} & \textbf{Other} & \textbf{Overall} \\
            \midrule
            \multicolumn{7}{c}{\textbf{Gen. Only}}                                                                                                                  \\
            \midrule
            SDv1.5~\citep{rombach2022high}         & 74.63           & 74.23           & 75.39              & 73.49             & 67.81          & 63.18            \\
            SD3-Medium~\citep{esser2024scaling}    & 87.90           & 91.01           & 88.83              & 80.70             & 88.68          & 84.08            \\
            SDXL~\citep{podell2024sdxl}            & 83.27           & 82.43           & 80.91              & 86.76             & 80.41          & 74.65            \\
            PixArt-$\alpha$~\citep{chen2024pixart} & 74.97           & 79.32           & 78.60              & 82.57             & 76.96          & 71.11            \\
            FLUX.1-dev~\citep{fluxdev2024}         & 74.35           & 90.00           & 88.96              & 90.87             & 88.33          & 83.84            \\
            \midrule
            \multicolumn{7}{c}{\textbf{Unified Models}}                                                                                                             \\
            \midrule
            Janus~\citep{wu2024janus}              & 82.33           & 87.38           & 87.70              & 85.46             & 86.41          & 79.68            \\
            Janus-Pro-1B~\citep{chen2025januspro}  & 87.58           & 88.63           & 88.17              & 88.98             & 88.30          & 82.63            \\
            Janus-Pro-7B~\citep{chen2025januspro}  & 86.90           & 88.90           & 89.40              & 89.32             & 89.48          & 84.19            \\
            MetaQuery-B~\citep{pan2025transfer}    & -               & -               & -                  & -                 & -              & 80.04            \\
            MetaQuery-L~\citep{pan2025transfer}    & -               & -               & -                  & -                 & -              & 81.10            \\
            MetaQuery-XL~\citep{pan2025transfer}   & -               & -               & -                  & -                 & -              & 82.05            \\
            BLIP3-o-4B~\citep{chen2025blipo}       & -               & -               & -                  & -                 & -              & 79.36            \\
            BLIP3-o-8B*~\citep{chen2025blipo}      & -               & -               & -                  & -                 & -              & 81.60            \\

            \midrule
            \multicolumn{7}{c}{\textbf{Ours}}                                                                                                                       \\
            \midrule
            \method-B 512                          & 91.33           & 89.39           & 90.07              & 86.89             & 87.78          & 82.95            \\
            \method-B 1024                         & 86.20           & 88.39           & 87.69              & 90.11             & 87.05          & 80.71            \\
            \method-L 512                          & 83.28           & 83.61           & 84.69              & 83.46             & 79.83          & 76.09            \\
            \method-L 1024                         & 79.27           & 82.00           & 80.93              & 82.81             & 78.68          & 72.26            \\
            \bottomrule
        \end{tabular}%
    \vspace{-1em}
\end{table*}

\subsection{WISE Evaluation Results}
The \tabref{tab:wise_benchmark} shows the detailed results of the WISE evaluation shown in \tabref{tab:geneval_benchmark}.
\begin{table*}[h]
    \centering
    \caption{\small{\textbf{WISE evaluation results.} Here BLIP3-o-8B* donates the model that is trained with an 30 million proprietary data.}}
    \vspace{-0.5em}
    \label{tab:wise_benchmark}
        \begin{tabular}{lccccccc}
            \toprule
            \textbf{METHOD}                        & \textbf{Cultural} & \textbf{Time} & \textbf{Space} & \textbf{Biology} & \textbf{Physics} & \textbf{Chemistry} & \textbf{Overall} \\
            \midrule
            \multicolumn{8}{c}{\textbf{Gen. Only}}                                                                                                                                    \\
            \midrule
            SDv1.5~\citep{rombach2022high}         & 0.34              & 0.35          & 0.32           & 0.28             & 0.29             & 0.21               & 0.32             \\
            SDv2.1~\citep{rombach2022high}         & 0.30              & 0.38          & 0.35           & 0.33             & 0.34             & 0.21               & 0.32             \\
            SD3-Medium~\citep{esser2024scaling}    & 0.42              & 0.44          & 0.48           & 0.39             & 0.47             & 0.29               & 0.42             \\
            SDXL~\citep{podell2024sdxl}            & 0.43              & 0.48          & 0.47           & 0.44             & 0.45             & 0.27               & 0.43             \\
            SD3.5-Large~\citep{esser2024scaling}   & 0.44              & 0.50          & 0.58           & 0.44             & 0.52             & 0.31               & 0.46             \\
            PixArt-$\alpha$~\citep{chen2024pixart} & 0.45              & 0.50          & 0.48           & 0.49             & 0.56             & 0.34               & 0.47             \\
            FLUX.1-dev~\citep{chen2024pixart}      & 0.48              & 0.58          & 0.62           & 0.42             & 0.51             & 0.35               & 0.50             \\
            \midrule
            \multicolumn{8}{c}{\textbf{Unified Models}}                                                                                                                               \\
            \midrule
            Show-o~\citep{xie2025showo}            & 0.28              & 0.40          & 0.48           & 0.30             & 0.46             & 0.30               & 0.35             \\
            Janus~\citep{wu2024janus}              & 0.16              & 0.26          & 0.35           & 0.28             & 0.30             & 0.14               & 0.23             \\
            Janus-Pro-1B~\citep{chen2025januspro}  & 0.20              & 0.28          & 0.45           & 0.24             & 0.32             & 0.16               & 0.26             \\
            Janus-Pro-7B~\citep{chen2025januspro}  & 0.30              & 0.37          & 0.49           & 0.36             & 0.42             & 0.26               & 0.35             \\
            MetaQuery-B~\citep{pan2025transfer}    & 0.44              & 0.49          & 0.58           & 0.41             & 0.49             & 0.34               & 0.46             \\
            MetaQuery-L~\citep{pan2025transfer}    & 0.56              & 0.57          & 0.62           & 0.48             & 0.63             & 0.42               & 0.55             \\
            MetaQuery-XL~\citep{pan2025transfer}   & 0.56              & 0.55          & 0.62           & 0.49             & 0.63             & 0.41               & 0.55             \\
            BAGEL~\citep{deng2025emerging}         & 0.44              & 0.55          & 0.68           & 0.44             & 0.60             & 0.39               & 0.52             \\
            BLIP3-o-4B~\citep{chen2025blipo}       & -                 & -             & -              & -                & -                & -                  & 0.50             \\
            BLIP3-o-8B*~\citep{chen2025blipo}      & -                 & -             & -              & -                & -                & -                  & 0.62             \\

            \midrule
            \multicolumn{8}{c}{\textbf{Ours}}                                                                                                           \\
            \midrule
            \method-B 512                          & 0.50              & 0.56          & 0.66           & 0.49             & 0.72             & 0.46               & 0.55             \\
            \method-B 1024                         & 0.44              & 0.50          & 0.64           & 0.46             & 0.63             & 0.43               & 0.50             \\
            \method-L 512                          & 0.48              & 0.56          & 0.63           & 0.49             & 0.64             & 0.51               & 0.53             \\
            \method-L 1024                         & 0.44              & 0.48          & 0.59           & 0.43             & 0.58             & 0.44               & 0.48             \\
            \bottomrule
        \end{tabular}%
    \vspace{-1em}
\end{table*}

\subsection{Different training recipe}
The results presented in \tabref{tab:different_training_recipe} correspond to the training configurations depicted in \figref{fig:training_recipe}. All models underwent approximately 5 epochs of pre-training on a dataset comprising 11 million images, followed by 10 epochs of fine-tuning on a dataset of approximately 210,000 images. Notably, the model pre-trained exclusively on image-only data and fine-tuned on a mixed data achieved superior performance across most metrics in the GenEval benchmark.
\begin{table*}[h]
    \centering
    \caption{\small{\textbf{Training recipe comparison.} The GenEval score of the models pre-trained with different training recipes. \textbf{Bold} denotes the best performance and \underline{underline} denotes the second best performance.}}
    \vspace{-0.5em}
    \label{tab:different_training_recipe}
    \resizebox{\textwidth}{!}{%
        \begin{tabular}{l|ccccccc}
            \toprule
            \textbf{Finetuning Recipe} & \textbf{Single Obj.} & \textbf{Two Obj.} & \textbf{Counting} & \textbf{Colors}  & \textbf{Position} & \textbf{Color Attri.} & \textbf{Overall} ($\uparrow$) \\
            \midrule
            \multicolumn{8}{c}{\textbf{Pre-trained with Text-Image Pair Data}}                                                                                                                        \\
            \midrule
            Image                      & \textbf{1.00}        & \textbf{0.95}     & 0.63              & 0.87             & 0.50              & 0.72                  & 0.78                          \\
            Pair                       & 0.99                 & \underline{0.92}  & 0.76              & 0.91             & 0.87              & 0.69                  & 0.86                          \\
            Mix                        & 0.99                 & 0.91              & \underline{0.80}  & 0.92             & \underline{0.90}  & 0.75                  & \underline{0.88}              \\
            \midrule
            \multicolumn{8}{c}{\textbf{Pre-trained with Image-Only Data}}                                                                                                                             \\
            \midrule
            Image                      & 0.99                 & 0.84              & 0.24              & 0.75             & 0.37              & 0.45                  & 0.61                          \\
            Pair                       & 0.99                 & 0.91              & 0.77              & \underline{0.93} & 0.87              & 0.75                  & 0.87                          \\
            Mix                        & 0.99                 & \underline{0.92}  & \textbf{0.83}     & \textbf{0.94}    & \textbf{0.91}     & 0.75                  & \textbf{0.89}                 \\
            \bottomrule
        \end{tabular}%
    }
    \vspace{-1em}
\end{table*}

\subsection{Image Editing Results}
\label{app:image_edit}
\begin{figure}[h]
    \centering
    \includegraphics[width=\linewidth]{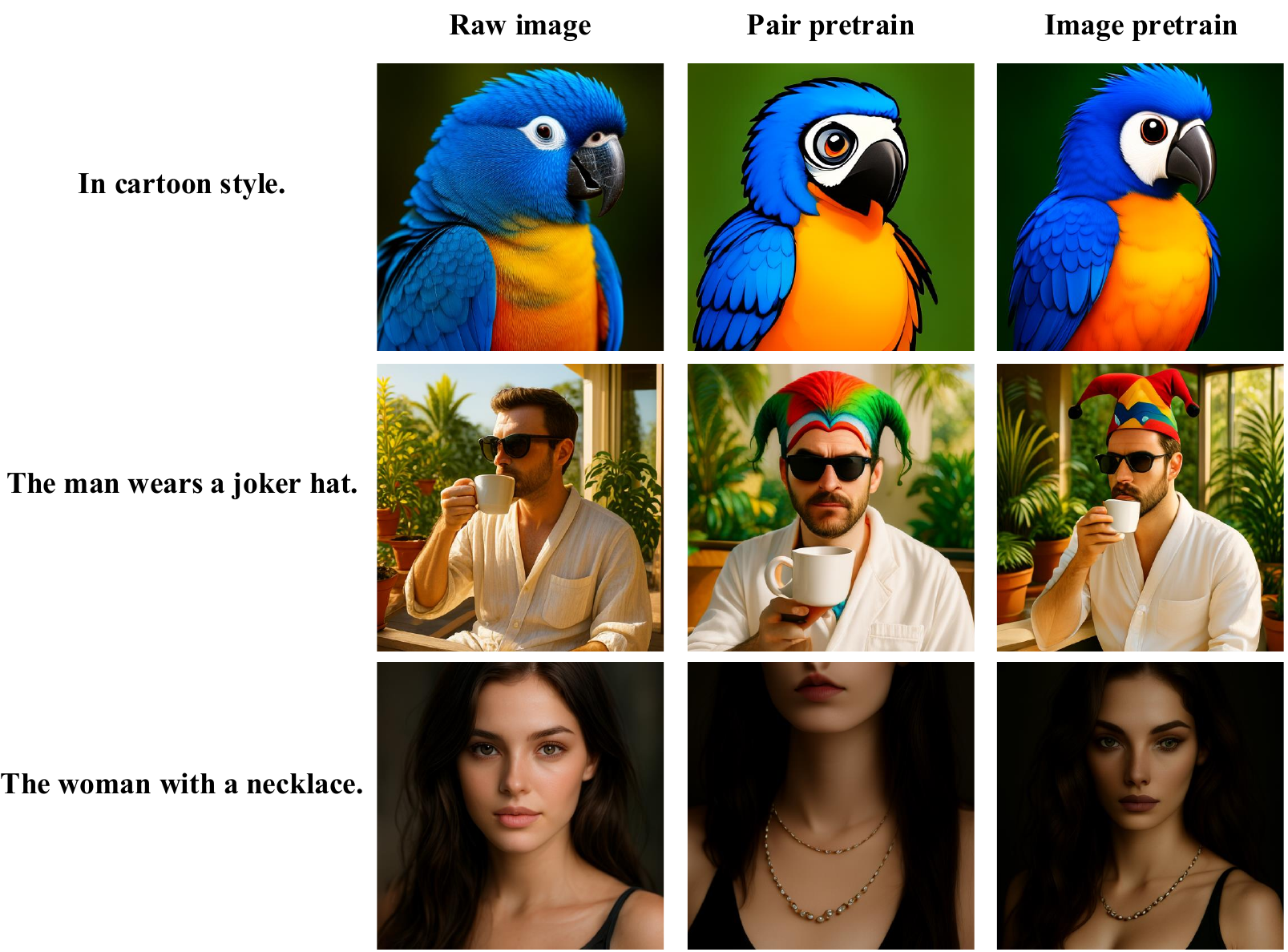}
    \caption{\textbf{Image editing ability with different pre-training method.}}
    \label{fig:image_edit}

\end{figure}

\figref{fig:image_edit} compares the image editing capabilities of models pre-trained exclusively on image-only data (right) and those pre-trained on image-text pairs (middle). The sole distinction between these models lies in their pre-training data type; all other hyperparameters and fine-tuning settings remain consistent. Despite in a \textit{zero-shot setting}, the model pre-trained with image-only data demonstrates superior consistency with the original input image. For instance, in the first row, the right image closely resembles the raw input, while in the second and third rows, the right images maintain nearly identical gestures to the original.

\subsection{UMM finetune result}
\tabref{tab:umm_finetuning_wise} show the detailed WISE score of the UMM finetuning results shown in \tabref{tab:umm_finetuning}.

\begin{table*}[h]
    \centering
    \caption{\small{\textbf{UMM finetuning WISE results.} Notation A\rp B denotes the result obtained by combining methods A and B.}}
    \vspace{-0.5em}
    \label{tab:umm_finetuning_wise}
    \resizebox{\textwidth}{!}{%
        \begin{tabular}{lcc|ccccccc}
            \toprule
            \textbf{METHOD}                                   & \textbf{Res.} & \textbf{NFEs}    & \textbf{Cultural} & \textbf{Time} & \textbf{Space} & \textbf{Biology} & \textbf{Physics} & \textbf{Chemistry} & \textbf{Overall} \\
            \midrule
            OpenUni-L~\citep{wu2025openuni}                   & 512           & 20$\times$2      & 0.51              & 0.45          & 0.58           & 0.39             & 0.50             & 0.30               & 0.52             \\
            \fd{\quad \rp Image finetuning}                   & \fd{512}      & \fd{20$\times$2} & \fd{0.46}         & \fd{0.52}     & \fd{0.66}      & \fd{0.49}        & \fd{0.51}        & \fd{0.29}          & \fd{0.49}        \\
            \quad \rp Pair finetuning                         & 512           & 20$\times$2      & 0.63              & 0.58          & 0.74           & 0.57             & 0.71             & 0.44               & 0.62             \\
            \quad \rp Mix finetuning                          & 512           & 20$\times$2      & 0.60              & 0.58          & 0.70           & 0.51             & 0.64             & 0.46               & 0.59             \\
            \midrule
            Qwen-Image~\citep{wu2025qwenimagetechnicalreport} & 512           & 50$\times$2      & -                 & -             & -              & -                & -                & -                  & -                \\
            \fd{\quad \rp Image finetuning}                   & \fd{512}      & \fd{50$\times$2} & \fd{0.39}         & \fd{0.42}     & \fd{0.56}      & \fd{0.32}        & \fd{0.50}        & \fd{0.28}          & \fd{0.41}        \\
            \quad \rp Pair finetuning                         & 512           & 50$\times$2      & 0.62              & 0.62          & 0.76           & 0.56             & 0.74             & 0.36               & 0.62             \\
            \quad \rp Mix finetuning                          & 512           & 50$\times$2      & 0.62              & 0.64          & 0.81           & 0.56             & 0.70             & 0.36               & 0.63             \\
            \midrule
            Qwen-Image~\citep{wu2025qwenimagetechnicalreport} & 1024          & 50$\times$2      & 0.62              & 0.63          & 0.77           & 0.57             & 0.75             & 0.40               & 0.62             \\
            \fd{\quad \rp Image finetuning}                   & \fd{1024}     & \fd{50$\times$2} & \fd{0.28}         & \fd{0.35}     & \fd{0.52}      & \fd{0.40}        & \fd{0.40}        & \fd{0.28}          & \fd{0.35}        \\
            \quad \rp Pair finetuning                         & 1024          & 50$\times$2      & 0.63              & 0.63          & 0.77           & 0.62             & 0.72             & 0.37               & 0.63             \\
            \quad \rp Mix finetuning                          & 1024          & 50$\times$2      & 0.64              & 0.63          & 0.78           & 0.57             & 0.73             & 0.38               & 0.63             \\
            \bottomrule
        \end{tabular}%
    }
    \vspace{-1em}
\end{table*}

\subsection{Generation results comparison of UMM finetuning}
\label{app:umm_finetune_result}
\begin{figure}[h]
    \centering
    \includegraphics[width=\linewidth]{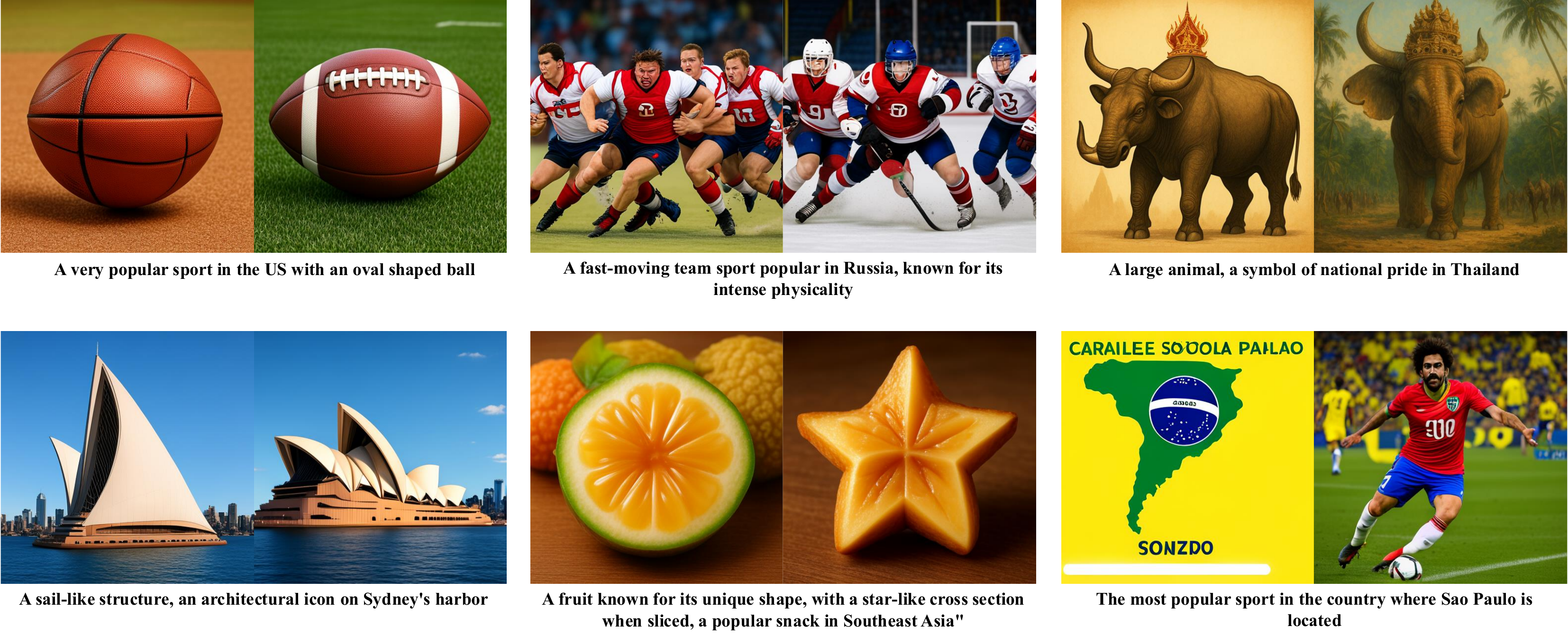}
    \caption{\textbf{Generation results of OpenUni-L before and after finetuning.} The left one is the image generated by the original OpenUni-L, while the right one is generated by the OpenUni-L after finetuning.}
    \label{fig:umm_generation_results}
\end{figure}

As illustrated in \figref{fig:umm_generation_results}, fine-tuning enhances the model's performance on tasks requiring reasoning. Although the understanding module was frozen during fine-tuning, the model's improved alignment between images and text enables more accurate generation of desired details.
What's more, a qualitative comparison between the original Qwen-Image model and our fine-tuned version. Our method enhances the model's ability to generate images with richer visual detail and improved alignment to the textual prompt.

\begin{figure*}[!t]
    \centering
    \begin{subfigure}[b]{0.49\textwidth}
        \centering
        \includegraphics[width=\linewidth]{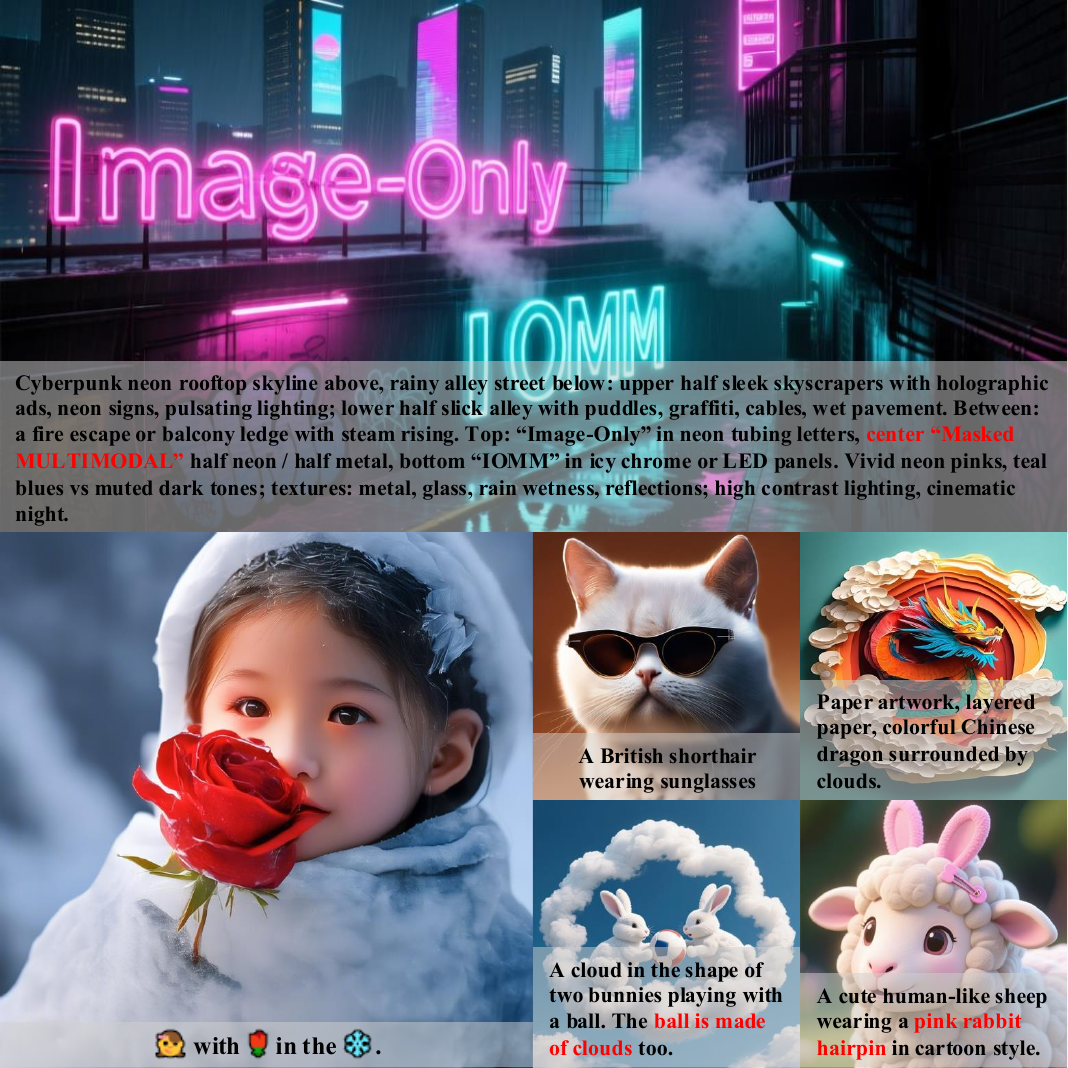}
        \caption{\textbf{Baseline Qwen-Image generation.}}
        \label{fig:raw_qwen_image_generation}
    \end{subfigure}
    \hfill
    \begin{subfigure}[b]{0.49\textwidth}
        \centering
        \includegraphics[width=\linewidth]{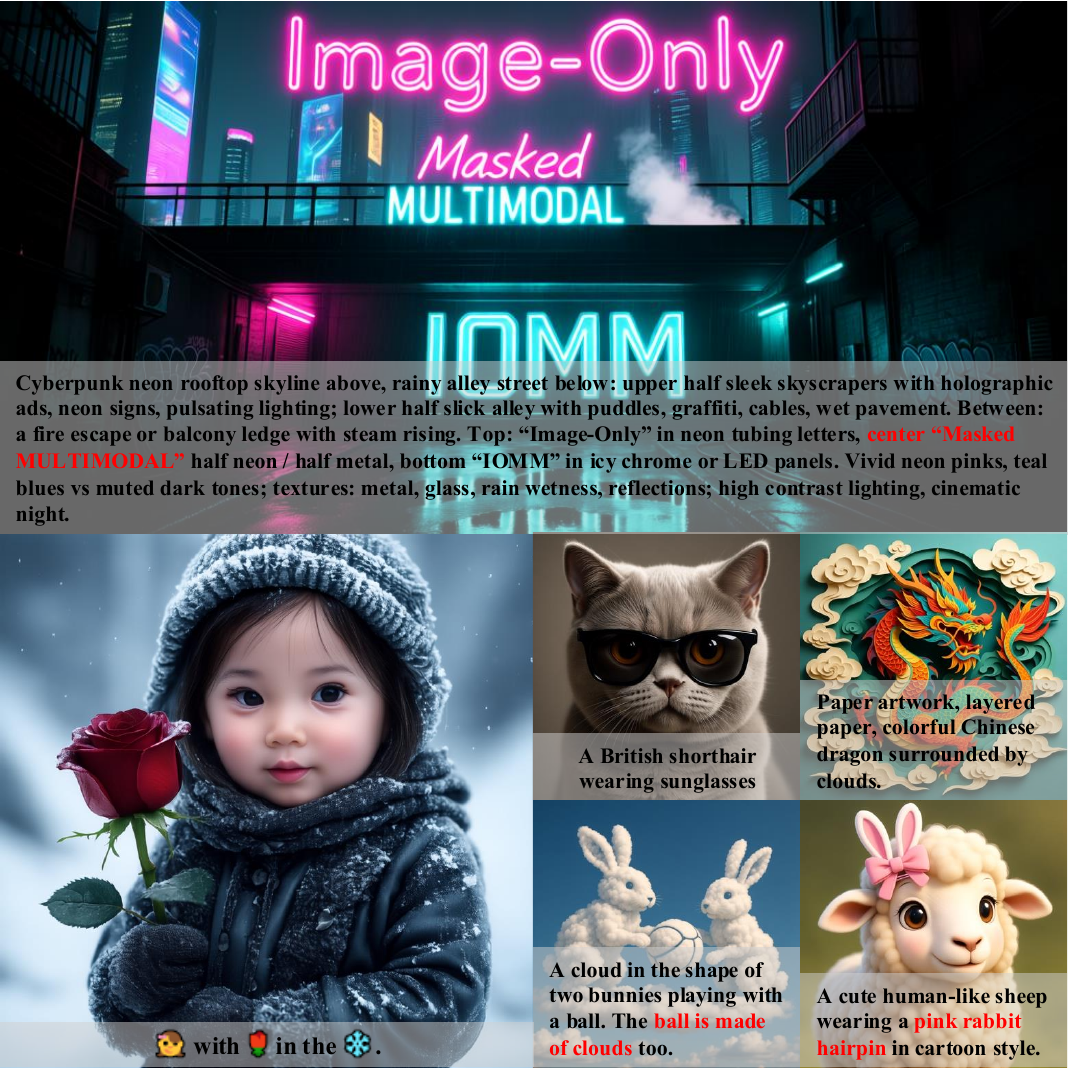}
        \caption{\textbf{Our fine-tuned Qwen-Image generation.}}
        \label{fig:new_qwen_image_generation}
    \end{subfigure}
    \vspace{-1em}
    \caption{\small{
            \textbf{(a, b)} Qualitative comparison between the original Qwen-Image model and our fine-tuned version. Our method enhances the model's ability to generate images with richer visual detail and improved alignment to the textual prompt.
        }}
    \vspace{-1em}
\end{figure*}

\newpage
\subsection{Prompts details}
\label{app:prompts_details}
The prompts used in \figref{fig:main_image} are as follows, from left to right, top to bottom.
\begin{itemize}
    \item Hyper-detailed macro photograph of a mechanical hummingbird crafted from gold filigree and sapphire gears, sipping nectar from a chrome rose; studio lighting, 200 mm macro lens, razor-sharp focus with creamy bokeh.
    \item A photo of a bear made entirely of autumn leaves.
    \item A fox wearing a suit and tie reading a newspaper at a café.
    \item a tiny astronaut hatching from an egg on the moon
    \item A man sipping coffee on a sunny balcony filled with potted plants, wearing linen clothes and sunglasses, basking in the morning light.
    \item A cloud in the shape of two bunnies playing with a ball. The ball is made of clouds too.
    \item Portrait of a noble samurai android wearing lacquered carbon-fiber armor and cherry-blossom patterns; Rembrandt lighting, 50 mm f/1.2, hyperreal pores and brushed metal textures.
    \item A hot air balloon in the shape of a heart. Grand Canyon
    \item A captivating photograph of an exquisite wooden dragon sculpture, skillfully carved with intricate details and realistic scales. The dragon is poised on a tree branch, its grand wings spread wide, revealing a mesmerizing woodland landscape below. The sky is painted with a symphony of soft blues and yellows, as the sun casts its final rays beyond the horizon. The dragon's glass eyes lend it a lifelike presence.
    \item Close-up portrait of a young woman with light skin and long brown hair, looking directly at the camera. Her face is illuminated by dramatic, slatted sunlight casting shadows across her features, creating a pattern of light and shadow. Her eyes are a striking green, and her lips are slightly parted, with a natural pink hue. The background is a soft, dark gradient, enhancing the focus on her face. The lighting is warm and golden.
    \item A lone figure in dark robes ascends worn stone steps toward a glowing light in an ancient temple entrance. Ornate arches, lush greenery, and intricate carvings adorn the scene, evoking a mystical, high-fantasy atmosphere reminiscent of works by artists like Randy Vargas, with cinematic lighting and epic storytelling.
    \item A whimsical scene featuring a plush toy bear wearing a blue sweater, positioned in the foreground, holding a butterfly on its raised arm. The bear is surrounded by a field of vibrant blue flowers, likely nemophila, creating a lush and colorful foreground. In the background, Mount Fuji rises majestically, its snow-capped peak sharply contrasting against a clear blue sky. The mountain is framed by fluffy white clouds and a line of dark green trees at its base. The butterfly, with its intricate black and orange wings, adds a touch of realism to the playful composition.
    \item A candid midday portrait of a young East Asian woman with dark braided hair, laughing softly at the camera while cradling a steaming mug of coffee. She wears a tattered band t-shirt with a faded punk logo, frayed gray collar, and missing sleeve button. The background shows peeling floral wallpaper and a rusted folding chair beneath a window with harsh noon sunlight. Shot as a grainy film photograph with high contrast and sharp focus on her animated expression.
    \item professional portrait photo of an anthropomorphic cat wearing fancy gentleman hat and jacket walking in autumn forest.
\end{itemize}

%% file: resources/reference.bib
@article{paszke2019pytorch,
  author  = {Paszke, A},
  journal = {arXiv preprint arXiv:1912.01703},
  title   = {Pytorch: An imperative style, high-performance deep learning library},
  year    = {2019}
}

@article{loshchilov2017decoupled,
  author  = {Loshchilov, Ilya and Hutter, Frank},
  journal = {arXiv preprint arXiv:1711.05101},
  title   = {Decoupled weight decay regularization},
  year    = {2017}
}

@inproceedings{rombach2022high,
  author    = {Rombach, Robin and Blattmann, Andreas and Lorenz, Dominik and Esser, Patrick and Ommer, Bj{\"o}rn},
  booktitle = {Proceedings of the IEEE/CVF conference on computer vision and pattern recognition},
  pages     = {10684--10695},
  title     = {High-resolution image synthesis with latent diffusion models},
  year      = {2022}
}

@article{vaswani2017attention,
  title   = {Attention is all you need},
  author  = {Vaswani, Ashish and Shazeer, Noam and Parmar, Niki and Uszkoreit, Jakob and Jones, Llion and Gomez, Aidan N and Kaiser, {\L}ukasz and Polosukhin, Illia},
  journal = {Advances in neural information processing systems},
  volume  = {30},
  year    = {2017}
}

@article{xie2024sana,
  author  = {Xie, Enze and Chen, Junsong and Chen, Junyu and Cai, Han and Tang, Haotian and Lin, Yujun and Zhang, Zhekai and Li, Muyang and Zhu, Ligeng and Lu, Yao and others},
  journal = {arXiv preprint arXiv:2410.10629},
  title   = {Sana: Efficient high-resolution image synthesis with linear diffusion transformers},
  year    = {2024}
}

@article{chen2025sana,
  author  = {Chen, Junsong and Xue, Shuchen and Zhao, Yuyang and Yu, Jincheng and Paul, Sayak and Chen, Junyu and Cai, Han and Xie, Enze and Han, Song},
  journal = {arXiv preprint arXiv:2503.09641},
  title   = {SANA-Sprint: One-Step Diffusion with Continuous-Time Consistency Distillation},
  year    = {2025}
}

@article{song2020score,
  author  = {Song, Yang and Sohl-Dickstein, Jascha and Kingma, Diederik P and Kumar, Abhishek and Ermon, Stefano and Poole, Ben},
  journal = {arXiv preprint arXiv:2011.13456},
  title   = {Score-based generative modeling through stochastic differential equations},
  year    = {2020}
}

@inproceedings{esser2024scaling,
  author    = {Esser, Patrick and Kulal, Sumith and Blattmann, Andreas and Entezari, Rahim and M{\"u}ller, Jonas and Saini, Harry and Levi, Yam and Lorenz, Dominik and Sauer, Axel and Boesel, Frederic and others},
  booktitle = {Forty-first international conference on machine learning},
  title     = {Scaling rectified flow transformers for high-resolution image synthesis},
  year      = {2024}
}

@article{lipman2022flow,
  author  = {Lipman, Yaron and Chen, Ricky TQ and Ben-Hamu, Heli and Nickel, Maximilian and Le, Matt},
  journal = {arXiv preprint arXiv:2210.02747},
  title   = {Flow matching for generative modeling},
  year    = {2022}
}

@inproceedings{ma2024sit,
  author       = {Ma, Nanye and Goldstein, Mark and Albergo, Michael S and Boffi, Nicholas M and Vanden-Eijnden, Eric and Xie, Saining},
  booktitle    = {European Conference on Computer Vision},
  organization = {Springer},
  pages        = {23--40},
  title        = {Sit: Exploring flow and diffusion-based generative models with scalable interpolant transformers},
  year         = {2024}
}

@article{yu2024representation,
  author  = {Yu, Sihyun and Kwak, Sangkyung and Jang, Huiwon and Jeong, Jongheon and Huang, Jonathan and Shin, Jinwoo and Xie, Saining},
  journal = {arXiv preprint arXiv:2410.06940},
  title   = {Representation alignment for generation: Training diffusion transformers is easier than you think},
  year    = {2024}
}

@article{sun2025unified,
  title   = {Unified Continuous Generative Models},
  author  = {Sun, Peng and Jiang, Yi and Lin, Tao},
  journal = {arXiv preprint arXiv:2505.07447},
  year    = {2025}
}

@article{ye2025echoo,
  title={Echo-4o: Harnessing the power of gpt-4o synthetic images for improved image generation},
  author={Ye, Junyan and Jiang, Dongzhi and Wang, Zihao and Zhu, Leqi and Hu, Zhenghao and Huang, Zilong and He, Jun and Yan, Zhiyuan and Yu, Jinghua and Li, Hongsheng and others},
  journal={arXiv preprint arXiv:2508.09987},
  year={2025}

}

@article{chen2025sharegptoimage,
  author    = {Chen, Junying and Cai, Zhenyang and Chen, Pengcheng and Chen, Shunian and Ji, Ke and Wang, Xidong and Yang, Yunjin and Wang, Benyou},
  journal   = {arXiv.org},
  year      = {2025},
  pages     = {},
  publisher = {},
  title     = {ShareGPT-4o-{Image}: Aligning {Multimodal} {Models} with {GPT}-4o-{Level} {Image} {Generation}},
  volume    = {abs/2506.18095}
}

@article{chen2025blipo,
  author    = {Chen, Jiuhai and Xu, Zhiyang and Pan, Xichen and Hu, Yushi and Qin, Can and Goldstein, Tom and Huang, Lifu and Zhou, Tianyi and Xie, Saining and Savarese, Silvio and Xue, Le and Xiong, Caiming and Xu, Ran},
  journal   = {arXiv.org},
  year      = {2025},
  pages     = {},
  publisher = {},
  title     = {BLIP3-o: A {Family} of {Fully} {Open} {Unified} {Multimodal} {Models}-{Architecture}, {Training} and {Dataset}},
  volume    = {abs/2505.09568}
}

@article{chen2025januspro,
  author    = {Chen, Xi-aokang and Wu, Zhiyu and Liu, Xingchao and Pan, Zizheng and Liu, Wen and Xie, Zhenda and Yu, Xingkai and Ruan, C.},
  journal   = {arXiv.org},
  year      = {2025},
  pages     = {},
  publisher = {},
  title     = {Janus-{Pro}: Unified {Multimodal} {Understanding} and {Generation} with {Data} and {Model} {Scaling}},
  volume    = {abs/2501.17811}
}

@inproceedings{ma2024janusflow,
  author       = {Ma, Yiyang and Liu, Xingchao and Chen, Xi-aokang and Liu, Wen and Wu, Chengyue and Wu, Zhiyu and Pan, Zizheng and Xie, Zhenda and Zhang, Haowei and Yu, Xingkai and Zhao, Liang and Wang, Yisong and Liu, Jiaying and Ruan, C.},
  booktitle    = {Computer {Vision} and {Pattern} {Recognition}},
  year         = {2024},
  pages        = {7739--7751},
  organization = {},
  title        = {JanusFlow: Harmonizing {Autoregression} and {Rectified} {Flow} for {Unified} {Multimodal} {Understanding} and {Generation}},
  volume       = {}
}

@inproceedings{wu2024janus,
  author       = {Wu, Chengyue and Chen, Xi-aokang and Wu, Zhiyu and Ma, Yiyang and Liu, Xingchao and Pan, Zizheng and Liu, Wen and Xie, Zhenda and Yu, Xingkai and Ruan, C. and Luo, Ping},
  booktitle    = {Computer {Vision} and {Pattern} {Recognition}},
  year         = {2024},
  pages        = {12966--12977},
  organization = {},
  title        = {Janus: Decoupling {Visual} {Encoding} for {Unified} {Multimodal} {Understanding} and {Generation}},
  volume       = {}
}

@article{pan2025transfer,
  author    = {Pan, Xichen and Shukla, Satya Narayan and Singh, Aashu and Zhao, Zhuokai and Mishra, Shlok Kumar and Wang, Jialiang and Xu, Zhiyang and Chen, Jiuhai and Li, Kunpeng and Juefei-Xu, Felix and Hou, Ji and Xie, Saining},
  journal   = {arXiv.org},
  year      = {2025},
  pages     = {},
  publisher = {},
  title     = {Transfer between {Modalities} with {MetaQueries}},
  volume    = {abs/2504.06256}
}

@inproceedings{xie2025showo,
  author       = {Xie, Jinheng and Mao, Weijia and Bai, Zechen and Zhang, David Junhao and Wang, Weihao and Lin, Kevin Qinghong and Gu, Yuchao and Chen, Zhijie and Yang, Zhenheng and Shou, Mike Zheng},
  booktitle    = {The {Thirteenth} {International} {Conference} on {Learning} {Representations}},
  year         = {2025},
  pages        = {},
  organization = {},
  title        = {Show-o: One {Single} {Transformer} to {Unify} {Multimodal} {Understanding} and {Generation}},
  volume       = {abs/2408.12528}
}

@inproceedings{zhou2025transfusion,
  author       = {Zhou, Chunting and YU, LILI and Babu, Arun and Tirumala, Kushal and Yasunaga, Michihiro and Shamis, Leonid and Kahn, Jacob and Ma, Xuezhe and Zettlemoyer, Luke and Levy, Omer},
  booktitle    = {The {Thirteenth} {International} {Conference} on {Learning} {Representations}},
  year         = {2025},
  pages        = {},
  organization = {},
  title        = {Transfusion: Predict the {Next} {Token} and {Diffuse} {Images} with {One} {Multi}-{Modal} {Model}},
  volume       = {abs/2408.11039}
}

@article{team2024chameleon,
  author    = {Team, Chameleon and Chen, Mingda and Kahn, Jacob},
  journal   = {arXiv.org},
  year      = {2024},
  pages     = {},
  publisher = {},
  title     = {Chameleon: Mixed-{Modal} {Early}-{Fusion} {Foundation} {Models}},
  volume    = {abs/2405.09818}
}

@inproceedings{podell2024sdxl,
  author       = {Podell, Dustin and English, Zion and Lacey, Kyle and Blattmann, A. and Dockhorn, Tim and Muller, Jonas and Penna, Joe and Rombach, Robin},
  booktitle    = {The {Twelfth} {International} {Conference} on {Learning} {Representations}},
  year         = {2024},
  pages        = {},
  organization = {},
  title        = {SDXL: Improving {Latent} {Diffusion} {Models} for {High}-{Resolution} {Image} {Synthesis}},
  volume       = {abs/2307.01952}
}

@inproceedings{chen2024pixart,
  author       = {Chen, Junsong and Yu, Jincheng and Ge, Chongjian and Yao, Lewei and Xie, Enze and Wu, Yue and Wang, Zhongdao and Kwok, James T. and Luo, Ping and Lu, Huchuan and Li, Zhenguo},
  booktitle    = {The {Twelfth} {International} {Conference} on {Learning} {Representations}},
  year         = {2024},
  pages        = {},
  organization = {},
  title        = {PixArt-$\alpha${}: Fast {Training} of {Diffusion} {Transformer} for {Photorealistic} {Text}-to-{Image} {Synthesis}},
  volume       = {abs/2310.00426}
}

@misc{text2image2m_2024,
  title        = {{text-to-image-2M}: A High-Quality, Diverse Text–Image Training Dataset},
  author       = {He, Jacky and contributors},
  year         = {2024},
  howpublished = {\url{https://huggingface.co/datasets/jackyhate/text-to-image-2M}},
  doi          = {10.57967/hf/3066}
}

@misc{matsubara2024megalith10m,
  title        = {{Megalith-10M}: A Dataset of 10 Million Public-Domain Photographs},
  author       = {Matsubara, Ollin and Draw Things AI Team},
  year         = {2024},
  howpublished = {\url{https://huggingface.co/datasets/madebyollin/megalith-10m}},
  note         = {CC0/Flickr-Commons images; Florence-2 captions available in the *megalith-10m-florence2* variant.}
}

@misc{labs2025flux1kontextflowmatching,
  title         = {FLUX.1 Kontext: Flow Matching for In-Context Image Generation and Editing in Latent Space},
  author        = {Black Forest Labs and Stephen Batifol and Andreas Blattmann and Frederic Boesel and Saksham Consul and Cyril Diagne and Tim Dockhorn and Jack English and Zion English and Patrick Esser and Sumith Kulal and Kyle Lacey and Yam Levi and Cheng Li and Dominik Lorenz and Jonas Müller and Dustin Podell and Robin Rombach and Harry Saini and Axel Sauer and Luke Smith},
  year          = {2025},
  eprint        = {2506.15742},
  archiveprefix = {arXiv},
  primaryclass  = {cs.GR},
  url           = {https://arxiv.org/abs/2506.15742}
}

@article{ramesh2022hierarchical,
  author    = {Ramesh, A. and Dhariwal, Prafulla and Nichol, Alex and Chu, Casey and Chen, Mark},
  journal   = {arXiv.org},
  year      = {2022},
  pages     = {},
  publisher = {},
  title     = {Hierarchical {Text}-{Conditional} {Image} {Generation} with {CLIP} {Latents}},
  volume    = {abs/2204.06125}
}

@article{dalle3,
  title   = {Improving image generation with better captions},
  author  = {Betker, James and Goh, Gabriel and Jing, Li and Brooks, Tim and Wang, Jianfeng and Li, Linjie and Ouyang, Long and Zhuang, Juntang and Lee, Joyce and Guo, Yufei and others},
  journal = {Computer Science. https://cdn. openai. com/papers/dall-e-3. pdf},
  volume  = {2},
  number  = {3},
  pages   = {8},
  year    = {2023}
}

@article{deng2025emerging,
  author    = {Deng, Chaorui and Zhu, Deyao and Li, Kunchang and Gou, Chenhui and Li, Feng and Wang, Zeyu and Zhong, Shu and Yu, Weihao and Nie, Xiaonan and Song, Ziang and Guang, Shi and Fan, Haoqi},
  journal   = {arXiv.org},
  year      = {2025},
  pages     = {},
  publisher = {},
  title     = {Emerging {Properties} in {Unified} {Multimodal} {Pretraining}},
  volume    = {abs/2505.14683}
}

@inproceedings{ghosh2023geneval,
  author       = {Ghosh, Dhruba and Hajishirzi, Hannaneh and Schmidt, Ludwig},
  booktitle    = {Conference on {Neural} {Information} {Processing} {Systems} ({NeurIPS})},
  year         = {2023},
  pages        = {},
  organization = {},
  title        = {GenEval: An object-focused framework for evaluating text-to-image alignment.},
  volume       = {}
}

@article{hu2024ella,
  author    = {Hu, Xiwei and Wang, Rui and Fang, Yixiao and Fu, Bin and Cheng, Pei and Yu, Gang},
  journal   = {arXiv.org},
  year      = {2024},
  pages     = {},
  publisher = {},
  title     = {ELLA: Equip {Diffusion} {Models} with {LLM} for {Enhanced} {Semantic} {Alignment}},
  volume    = {abs/2403.05135}
}

@article{ye2025imgedit,
  author    = {Ye, Yang and He, Xianyi and Li, Zongjian and Lin, Bin and Yuan, Shenghai and Yan, Zhiyuan and Hou, Bohan and Yuan, Li},
  journal   = {arXiv.org},
  year      = {2025},
  pages     = {},
  publisher = {},
  title     = {ImgEdit: A {Unified} {Image} {Editing} {Dataset} and {Benchmark}},
  volume    = {abs/2505.20275}
}

@article{wu2025openuni,
  author    = {Wu, Size and Wu, Zhonghua and Gong, Zerui and Tao, Qi and Jin, Sheng and Li, Qinyue and Li, Wei and Loy, Chen Change},
  journal   = {arXiv.org},
  year      = {2025},
  pages     = {},
  publisher = {},
  title     = {OpenUni: A {Simple} {Baseline} for {Unified} {Multimodal} {Understanding} and {Generation}},
  volume    = {abs/2505.23661}
}

@misc{wu2025qwenimagetechnicalreport,
  title         = {Qwen-Image Technical Report},
  author        = {Chenfei Wu and Jiahao Li and Jingren Zhou and Junyang Lin and Kaiyuan Gao and Kun Yan and Sheng-ming Yin and Shuai Bai and Xiao Xu and Yilei Chen and Yuxiang Chen and Zecheng Tang and Zekai Zhang and Zhengyi Wang and An Yang and Bowen Yu and Chen Cheng and Dayiheng Liu and Deqing Li and Hang Zhang and Hao Meng and Hu Wei and Jingyuan Ni and Kai Chen and Kuan Cao and Liang Peng and Lin Qu and Minggang Wu and Peng Wang and Shuting Yu and Tingkun Wen and Wensen Feng and Xiaoxiao Xu and Yi Wang and Yichang Zhang and Yongqiang Zhu and Yujia Wu and Yuxuan Cai and Zenan Liu},
  year          = {2025},
  eprint        = {2508.02324},
  archiveprefix = {arXiv},
  primaryclass  = {cs.CV},
  url           = {https://arxiv.org/abs/2508.02324}
}

@article{niu2025wise,
  author    = {Niu, Yuwei and Ning, Munan and Zheng, Mengren and Lin, Bin and Jin, Peng and Liao, Jiaqi and Ning, Kun-Peng and Zhu, Bin and Yuan, Li},
  journal   = {arXiv.org},
  year      = {2025},
  pages     = {},
  publisher = {},
  title     = {WISE: A {World} {Knowledge}-{Informed} {Semantic} {Evaluation} for {Text}-to-{Image} {Generation}},
  volume    = {abs/2503.07265}
}

@inproceedings{wu2025vilau,
  author       = {Wu, Yecheng and Zhang, Zhuoyang and Chen, Junyu and Tang, Haotian and Li, Dacheng and Fang, Yunhao and Zhu, Ligeng and Xie, Enze and Yin, Hongxu and Yi, Li and Han, Song and Lu, Yao},
  booktitle    = {The {Thirteenth} {International} {Conference} on {Learning} {Representations}},
  year         = {2025},
  pages        = {},
  organization = {},
  title        = {VILA-{U}: a {Unified} {Foundation} {Model} {Integrating} {Visual} {Understanding} and {Generation}},
  volume       = {abs/2409.04429}
}

@article{lin2025uniworldv,
  author    = {Lin, Bin and Li, Zongjian and Cheng, Xinhua and Niu, Yuwei and Ye, Yang and He, Xianyi and Yuan, Shenghai and Yu, Wangbo and Wang, Shaodong and Ge, Yunyang and Pang, Yatian and Yuan, Li},
  journal   = {arXiv.org},
  year      = {2025},
  pages     = {},
  publisher = {},
  title     = {UniWorld-{V1}: High-{Resolution} {Semantic} {Encoders} for {Unified} {Visual} {Understanding} and {Generation}},
  volume    = {abs/2506.03147}
}

@inproceedings{he2022masked,
  author       = {He, Kaiming and Chen, Xinlei and Xie, Saining and Li, Yanghao and Dollar, Piotr and Girshick, Ross},
  booktitle    = {2022 {IEEE}/{CVF} {Conference} on {Computer} {Vision} and {Pattern} {Recognition} ({CVPR})},
  year         = {2022},
  pages        = {},
  organization = {IEEE},
  title        = {Masked {Autoencoders} {Are} {Scalable} {Vision} {Learners}},
  volume       = {}
}

@article{zhang2023magicbrush,
  title   = {Magicbrush: A manually annotated dataset for instruction-guided image editing},
  author  = {Zhang, Kai and Mo, Lingbo and Chen, Wenhu and Sun, Huan and Su, Yu},
  journal = {Advances in Neural Information Processing Systems},
  volume  = {36},
  pages   = {31428--31449},
  year    = {2023}
}

@inproceedings{brooks2023instructpix2pix,
  title     = {Instructpix2pix: Learning to follow image editing instructions},
  author    = {Brooks, Tim and Holynski, Aleksander and Efros, Alexei A},
  booktitle = {Proceedings of the IEEE/CVF conference on computer vision and pattern recognition},
  pages     = {18392--18402},
  year      = {2023}
}

@inproceedings{yu2025anyedit,
  title     = {Anyedit: Mastering unified high-quality image editing for any idea},
  author    = {Yu, Qifan and Chow, Wei and Yue, Zhongqi and Pan, Kaihang and Wu, Yang and Wan, Xiaoyang and Li, Juncheng and Tang, Siliang and Zhang, Hanwang and Zhuang, Yueting},
  booktitle = {Proceedings of the Computer Vision and Pattern Recognition Conference},
  pages     = {26125--26135},
  year      = {2025}
}

@article{zhao2024ultraedit,
  title   = {Ultraedit: Instruction-based fine-grained image editing at scale},
  author  = {Zhao, Haozhe and Ma, Xiaojian Shawn and Chen, Liang and Si, Shuzheng and Wu, Rujie and An, Kaikai and Yu, Peiyu and Zhang, Minjia and Li, Qing and Chang, Baobao},
  journal = {Advances in Neural Information Processing Systems},
  volume  = {37},
  pages   = {3058--3093},
  year    = {2024}
}

@inproceedings{xiao2025omnigen,
  title     = {Omnigen: Unified image generation},
  author    = {Xiao, Shitao and Wang, Yueze and Zhou, Junjie and Yuan, Huaying and Xing, Xingrun and Yan, Ruiran and Li, Chaofan and Wang, Shuting and Huang, Tiejun and Liu, Zheng},
  booktitle = {Proceedings of the Computer Vision and Pattern Recognition Conference},
  pages     = {13294--13304},
  year      = {2025}
}

@article{zhang2025context,
  title   = {In-context edit: Enabling instructional image editing with in-context generation in large scale diffusion transformer},
  author  = {Zhang, Zechuan and Xie, Ji and Lu, Yu and Yang, Zongxin and Yang, Yi},
  journal = {arXiv preprint arXiv:2504.20690},
  year    = {2025}
}

@article{liu2025step1x,
  title   = {Step1x-edit: A practical framework for general image editing},
  author  = {Liu, Shiyu and Han, Yucheng and Xing, Peng and Yin, Fukun and Wang, Rui and Cheng, Wei and Liao, Jiaqi and Wang, Yingming and Fu, Honghao and Han, Chunrui and others},
  journal = {arXiv preprint arXiv:2504.17761},
  year    = {2025}
}

@article{labs2025kontext,
  title   = {FLUX. 1 Kontext: Flow Matching for In-Context Image Generation and Editing in Latent Space},
  author  = {Labs, Black Forest and Batifol, Stephen and Blattmann, Andreas and Boesel, Frederic and Consul, Saksham and Diagne, Cyril and Dockhorn, Tim and English, Jack and English, Zion and Esser, Patrick and others},
  journal = {arXiv preprint arXiv:2506.15742},
  year    = {2025}
}

@misc{playground-v2,
  url    = {[https://huggingface.co/playgroundai/playground-v2-1024px-aesthetic](https://huggingface.co/playgroundai/playground-v2-1024px-aesthetic)},
  title  = {Playground v2},
  author = {Li, Daiqing and Kamko, Aleks and Sabet, Ali and Akhgari, Ehsan and Xu, Linmiao and Doshi, Suhail}
}

@article{li2024playground,
  author    = {Li, Daiqing and Kamko, Aleks and Akhgari, Ehsan and Sabet, Ali and Xu, Linmiao and Doshi, Suhail},
  journal   = {arXiv.org},
  year      = {2024},
  pages     = {},
  publisher = {},
  title     = {Playground v2.5: Three {Insights} towards {Enhancing} {Aesthetic} {Quality} in {Text}-to-{Image} {Generation}},
  volume    = {abs/2402.17245}
}

@inproceedings{chen2024pixarts,
  author       = {Chen, Junsong and Ge, Chongjian and Xie, Enze and Wu, Yue and Yao, Lewei and Ren, Xiaozhe and Wang, Zhongdao and Luo, Ping and Lu, Huchuan and Li, Zhenguo},
  booktitle    = {European {Conference} on {Computer} {Vision} ({ECCV})},
  year         = {2024},
  pages        = {74--91},
  organization = {},
  title        = {PIXART-$\sigma$: Weak-to-{Strong} {Training} of {Diffusion} {Transformer} for 4K {Text}-to-{Image} {Generation}.},
  volume       = {}
}

@article{wu2025harmonizing,
  author    = {Wu, Size and Zhang, Wenwei and Xu, Lumin and Jin, Sheng and Wu, Zhonghua and Tao, Qi and Liu, Wentao and Li, Wei and Loy, Chen Change},
  journal   = {arXiv.org},
  year      = {2025},
  pages     = {},
  publisher = {},
  title     = {Harmonizing {Visual} {Representations} for {Unified} {Multimodal} {Understanding} and {Generation}},
  volume    = {abs/2503.21979}
}

@inproceedings{dong2024dreamllm,
  author       = {Dong, Runpei and Han, Chunrui and Peng, Yuang and Qi, Zekun and Ge, Zheng and Yang, Jinrong and Zhao, Liang and Sun, Jianjian and Zhou, Hongyu and Wei, Haoran and Kong, Xiangwen and Zhang, Xiangyu and Ma, Kaisheng and Yi, Li},
  booktitle    = {The {Twelfth} {International} {Conference} on {Learning} {Representations}},
  year         = {2024},
  pages        = {},
  organization = {},
  title        = {DreamLLM: Synergistic {Multimodal} {Comprehension} and {Creation}},
  volume       = {}
}

@article{xie2025showo2,
  author    = {Xie, Jinheng and Yang, Zhenheng and Shou, Mike Zheng},
  journal   = {arXiv.org},
  year      = {2025},
  pages     = {},
  publisher = {},
  title     = {Show-o2: Improved {Native} {Unified} {Multimodal} {Models}},
  volume    = {abs/2506.15564}
}

@online{Google2025Gemini,
  author  = {{Google}},
  title   = {Experiment with Gemini 2.0 Flash Native Image Generation},
  year    = {2025},
  url     = {https://developers.googleblog.com/en/experiment-with-gemini-20-flash-native-image-generation/},
  urldate = {2025-09-22}
}

@online{OpenAI2025Introducing,
  author  = {{OpenAI}},
  title   = {Introducing 4o Image Generation},
  year    = {2025},
  url     = {https://openai.com/index/introducing-4o-image-generation/},
  urldate = {2025-09-22}
}

@online{Google2025Gemini25FlashImage,
  author  = {{Google}},
  title   = {Gemini 2.5 Flash Image: High-Consistency Image Generation and Editing},
  year    = {2025},
  month   = {8},
  url     = {https://aistudio.google.com/models/gemini-2-5-flash-image},
  urldate = {2025-09-22},
  note    = {Official model page on Google AI Studio. Internal development code name: nano-banana.}
}

@inproceedings{chang2022maskgit,
  author       = {Chang, Huiwen and Zhang, Han and Jiang, Lu and Liu, Ce and Freeman, William T.},
  booktitle    = {Computer {Vision} and {Pattern} {Recognition} ({CVPR})},
  year         = {2022},
  pages        = {11305--11315},
  organization = {},
  title        = {MaskGIT: Masked {Generative} {Image} {Transformer}.},
  volume       = {}
}

@article{zhou2023maskdiffusion,
  author    = {Zhou, Yupeng and Zhou, Daquan and Zhu, Zuo-Liang and Wang, Yaxing and Hou, Qibin and Feng, Jiashi},
  journal   = {International Journal of Computer Vision},
  year      = {2023},
  pages     = {},
  publisher = {},
  title     = {MaskDiffusion: Boosting {Text}-to-{Image} {Consistency} with {Conditional} {Mask}},
  volume    = {abs/2309.04399}
}

@inproceedings{zou2024towards,
  author       = {Zou, Siyu and Tang, Jiji and Zhou, Yiyi and He, Jing and Zhao, Chaoyi and Zhang, Rongsheng and Hu, Zhipeng and Sun, Xiaoshuai},
  booktitle    = {AAAI {Conference} on {Artificial} {Intelligence} ({AAAI})},
  year         = {2024},
  pages        = {7864--7872},
  organization = {},
  title        = {Towards {Efficient} {Diffusion}-{Based} {Image} {Editing} with {Instant} {Attention} {Masks}.},
  volume       = {}
}

@inproceedings{huang2022masked,
  author       = {Huang, Jiaxing and Cui, Kaiwen and Guan, Dayan and Xiao, Aoran and Zhan, Fangneng and Lu, Shijian and Liao, Shengcai and Xing, Eric P.},
  booktitle    = {Conference on {Neural} {Information} {Processing} {Systems} ({NeurIPS})},
  year         = {2022},
  pages        = {},
  organization = {},
  title        = {Masked {Generative} {Adversarial} {Networks} are {Data}-{Efficient} {Generation} {Learners}.},
  volume       = {}
}

@misc{fluxdev2024,
  author       = {BlackForest},
  title        = {FLUX},
  year         = {2024},
  howpublished = {\url{https://github.com/black-forest-labs/flux}}
}

@article{zhu2025internvl3,
  title   = {Internvl3: Exploring advanced training and test-time recipes for open-source multimodal models},
  author  = {Zhu, Jinguo and Wang, Weiyun and Chen, Zhe and Liu, Zhaoyang and Ye, Shenglong and Gu, Lixin and Tian, Hao and Duan, Yuchen and Su, Weijie and Shao, Jie and others},
  journal = {arXiv preprint arXiv:2504.10479},
  year    = {2025}
}

@article{hu2022lora,
  title   = {Lora: Low-rank adaptation of large language models.},
  author  = {Hu, Edward J and Shen, Yelong and Wallis, Phillip and Allen-Zhu, Zeyuan and Li, Yuanzhi and Wang, Shean and Wang, Lu and Chen, Weizhu and others},
  journal = {ICLR},
  volume  = {1},
  number  = {2},
  pages   = {3},
  year    = {2022}
}

@inproceedings{ma2025learning,
  title     = {Learning visual generative priors without text},
  author    = {Ma, Shuailei and Zheng, Kecheng and Wei, Ying and Wu, Wei and Lu, Fan and Zhang, Yifei and Xie, Chen-Wei and Gong, Biao and Zhu, Jiapeng and Shen, Yujun},
  booktitle = {Proceedings of the Computer Vision and Pattern Recognition Conference},
  pages     = {8051--8061},
  year      = {2025}
}

@misc{yan2025unified,
  title         = {Unified Multimodal Model as Auto-Encoder},
  author        = {Zhiyuan Yan and Kaiqing Lin and Zongjian Li and Junyan Ye and Hui Han and Zhendong Wang and Hao Liu and Boyang Lin and Hui Li and Xiaodan Xu and Xin Xiao},
  year          = {2025},
  eprint        = {2509.09666},
  archiveprefix = {arXiv},
  primaryclass  = {cs.CV}
}

@article{xie2025reconstruction,
  title         = {{Reconstruction alignment improves unified multimodal models}},
  author        = {Xie, Ji and Darrell, Trevor and Zettlemoyer, Luke and Wang, XuDong},
  journal       = {arXiv preprint arXiv:2509.07295},
  year          = {2025},
  eprint        = {2509.07295},
  archiveprefix = {arXiv}
}

@inproceedings{wang2025visual,
  author    = {Wang, XuDong and Zhou, Xingyi and Fathi, Alireza and Darrell, Trevor and Schmid, Cordelia},
  title     = {{Visual lexicon: Rich image features in language space}},
  booktitle = {Proceedings of the Computer Vision and Pattern Recognition Conference},
  pages     = {19736--19747},
  year      = {2025}
}

@article{cai2025z,
  title={Z-image: An efficient image generation foundation model with single-stream diffusion transformer},
  author={Cai, Huanqia and Cao, Sihan and Du, Ruoyi and Gao, Peng and Hoi, Steven and Hou, Zhaohui and Huang, Shijie and Jiang, Dengyang and Jin, Xin and Li, Liangchen and others},
  journal={arXiv preprint arXiv:2511.22699},
  year={2025}
}

@article{jordan2024muon,
  title={Muon: An optimizer for hidden layers in neural networks, 2024},
  author={Jordan, Keller and Jin, Yuchen and Boza, Vlado and Jiacheng, You and Cesista, Franz and Newhouse, Laker and Bernstein, Jeremy},
  journal={URL https://kellerjordan. github. io/posts/muon},
  volume={6},
  number={3},
  pages={4},
  year={2024}
}
